\documentclass{article} % For LaTeX2e
\usepackage{iclr2025_conference,times}

\usepackage{amsmath,amsfonts,bm}

\def\eqref#1{equation~\ref{#1}}
\def\1{\bm{1}}

\def\mL{{\bm{L}}}

\DeclareMathAlphabet{\mathsfit}{\encodingdefault}{\sfdefault}{m}{sl}
\SetMathAlphabet{\mathsfit}{bold}{\encodingdefault}{\sfdefault}{bx}{n}

\usepackage{hyperref}
\usepackage{url}
\usepackage{bm}
\usepackage{microtype}
\usepackage{subfigure}
\usepackage{booktabs}
\usepackage{hyperref}
\usepackage{setspace}
\usepackage{bmpsize}
\usepackage{algorithm}
\usepackage{algpseudocode}
\definecolor{cvprblue}{rgb}{0.21,0.49,0.74}
\hypersetup{
    breaklinks=true,
    colorlinks=true,
    citecolor=cvprblue,
    linkcolor=purple,
    urlcolor=cyan,
}

\newcommand\nnfootnote[1]{%
  \renewcommand\thefootnote{}\footnote{#1}%
  \addtocounter{footnote}{-1}%
}
\usepackage[utf8]{inputenc} % allow utf-8 input
\usepackage[T1]{fontenc}    % use 8-bit T1 fonts
\usepackage{hyperref}       % hyperlinks
\usepackage{url}            % simple URL typesetting
\usepackage{booktabs}       % professional-quality tables
\usepackage{amsfonts}       % blackboard math symbols
\usepackage{nicefrac}       % compact symbols for 1/2, etc.
\usepackage{microtype}      % microtypography
\usepackage{xcolor}         % colors

\usepackage{amsmath}
\usepackage{amssymb}
\usepackage{mathtools}
\usepackage{amsthm}
\usepackage{amsfonts}
\usepackage{verbatim}
\usepackage{multirow}
\usepackage{wrapfig}
\usepackage{float}
\usepackage[capitalize,noabbrev]{cleveref}

\theoremstyle{plain}
\newtheorem{theorem}{Theorem}[section]

\newtheorem{lemma}[theorem]{Lemma}

\theoremstyle{definition}
\newtheorem{definition}[theorem]{Definition}
\newtheorem{assumption}[theorem]{Assumption}
\theoremstyle{remark}

\usepackage[textsize=tiny]{todonotes}

\iclrfinalcopy 
\begin{document}

\title{Second-Order Fine-Tuning without Pain for LLMs: A Hessian Informed Zeroth-Order \\Optimizer}

\author{
Yanjun Zhao$^{1,*}$,~~Sizhe Dang$^{1,*}$,~~Haishan Ye$^{1,\dagger}$,~~Guang Dai$^2$,~~Yi Qian$^1$,~~Ivor W.Tsang$^3$\\
$^1$Xi'an Jiaotong University, China, $^2$SGIT AI Lab, Xi'an, China\\
$^3$Centre for Frontier Artificial Intelligence Research, A* STAR, Singapore\\
\texttt{\{yanjun.zhao, darknight1118\}@stu.xjtu.edu.cn}\\
\texttt{yehaishan@xjtu.edu.cn},~~
\texttt{yqian@mail.xjtu.edu.cn}\\
\texttt{guang.dai@gmail.com},~~
\texttt{ivor\_tsang@cfar.a-star.edu.sg}\\
}

% \customfootnotetext{1}{\textsuperscript{*}Equal contribution. \textsuperscript{\dagger}Correspondence author.}
\nnfootnote{$^*$ Equal contribution. $^\dagger$ Correspondence author}

\newcommand{\fix}{\marginpar{FIX}}
\newcommand{\new}{\marginpar{NEW}}

\maketitle

\newcommand{\name}{HiZOO}
\newcommand{\nameo}{HiZOO }
\newcommand{\cO}{\mathcal{O}}
\newcommand{\RR}{\mathbb{R}}
\newcommand{\EE}{\mathbb{E}}

\begin{abstract}
Fine-tuning large language models (LLMs) is necessary for specific downstream tasks, but the classic adaptive first-order optimizer entails prohibitive GPU memory because of backpropagation. Recent works such as MeZO have turned to zeroth-order optimizers for fine-tuning, which reduce substantial memory by using just two forward passes. However, heterogeneous curvatures across different parameter dimensions in LLMs often cause convergence instability or even failure. In this work, we propose HiZOO, a diagonal \textbf{H}essian \textbf{i}nformed \textbf{Z}eroth-\textbf{O}rder \textbf{O}ptimizer , which is the first to leverage the diagonal Hessian to enhance ZOO for fine-tuning LLMs. We provide the theoretical proof for \nameo and visualize the optimization trajectories on the test functions. %to illustrate how it improves convergence in handling heterogeneous curvatures. 
Extensive experiments on various models (RoBERTa, OPT, Phi-2, and LLama3, with 350M$\sim$66B parameters) indicate that \nameo significantly reduces the number of training steps and improves model accuracy. For example, on the SST2 task, HiZOO achieves an \textbf{8}\boldsymbol{$\times$} speed-up and better accuracy. Even when scaled to 66B-model, HiZOO outperforms MeZO with up to \textbf{5.1\%} absolute improvement. We also propose HiZOO-L, which reduces the Hessian memory cost to \textbf{10\%} of the MeZO, while maintaining almost same performance. Compared with ZO-Adam, HiZOO-L achieves a \textbf{4.3\%} absolute improvement, just using \textbf{50\%} of the GPU memory. Code is available at https://github.com/Yanjun-Zhao/HiZOO.
\end{abstract}

\begin{figure}[ht]
\vskip -0.2in
\centering
\scalebox{1.0}{
%,bb=0 0 2200 800]
\includegraphics[width=0.95\linewidth]{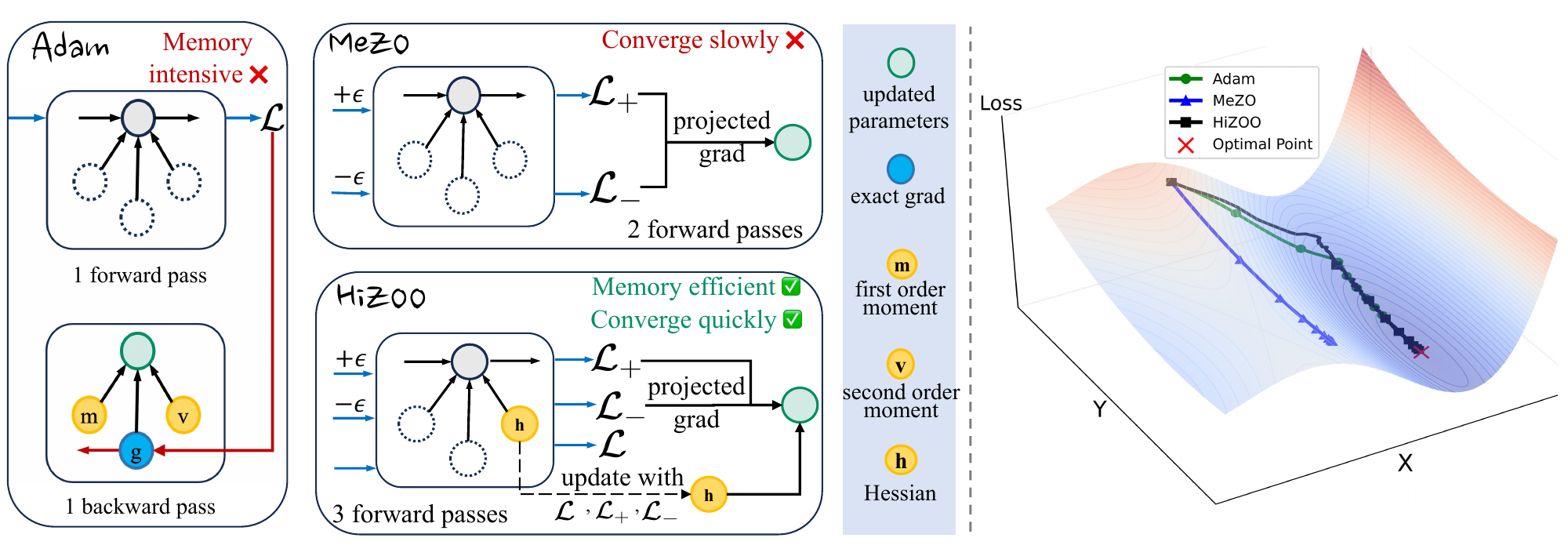}}
%\vspace{-3mm}
\vskip -0.15in
\caption{(Left) Comparison of HiZOO, MeZO and Adam. (Right) \textbf{Heterogeneous curvatures example}. \nameo updates along the direction with greater curvature ($X$) and converges more quickly than MeZO. The corresponding loss curves are shown in Section~\ref{2D_curvature}.}
\label{fig:3D}
\vskip -0.1in
\end{figure}

\vskip -0.2in
\section{Introduction}
\label{sec_introduction}

% \begin{figure}[ht]
% \vskip -0.2in
% \centering
% \scalebox{1.0}{
% %,bb=0 0 2200 800]
% \includegraphics[width=0.95\linewidth]{Figure/intro.pdf}}
% %\vspace{-3mm}
% \vskip -0.15in
% \caption{(Left) Comparison of HiZOO, MeZO and Adam. (Right) \textbf{Heterogeneous curvatures example}. \nameo updates along the direction with greater curvature ($X$) and converges more quickly than MeZO. The corresponding loss curves are shown in Section~\ref{2D_curvature}.}
% \label{fig:3D}
% \vskip -0.1in
% \end{figure}

Fine-tuning pre-trained LLMs for specific tasks has gained significant attention recently. As the number of model parameters increases, full parameter fine-tuning (FT) becomes markedly memory-intensive. To alleviate GPU memory limitations, parameter-efficient fine-tuning (PEFT) methods~\citep{hu2022lora,li-liang-2021-prefix, dettmers2023qlora, zhao2024galore, pan2024lisa} have been developed, which only fine-tune a small number of (extra) model parameters. As a result, they significantly reduce the computational and storage cost, while achieving performance comparable to a fully fine-tuned model.

Adaptive first-order optimizers such as Adam~\citep{Adam} and AdamW~\citep{adamw} are widely used to fine-tune LLMs. However, using these optimizers still leads to substantial memory consumption, primarily due to the inherent backpropagation process to calculate the gradient. To address these limitations, MeZO~\citep{mezo} proposed to utilize a zeroth-order optimizer (ZOO) to estimate the gradient with just two forward passes per step, no need for backpropagation anymore. This achieves numerous memory reductions and makes it accessible to train and store LLMs on consumer hardware.

However, the parameters of LLMs often exhibit heterogeneous curvatures across different dimensions, as documented in recent studies \citep{sagun2017eigenvalues, Hessian_Eigenvalue_Density, Adaptive_Methods_Good_for_Attention}. This significant difference of second derivative makes the MeZO converge towards saddle point, slowing down the convergence speed, as shown in Figure~\ref{fig:3D} (right). Since the incorporation of Hessian to measure the curvature properties of the loss landscape, second-order methods \citep{Liu2023sophia:, ADAHESSIAN, anil2021scalable} can solve this suboptimal behavior. Unfortunately, in the context of zeroth-order optimization, one cannot directly compute the Hessian atop first-order derivatives.

% \textcolor{blue}{First order methods only use gradient information and do not consider the curvature properties of the loss landscape, thereby leading to their sub-optimal behaviour. To solve this dilemma, the second order methods \citep{Liu2023sophia:, ADAHESSIAN, anil2021scalable} incorporate both gradient and Hessian. The main idea underlying these methods involves calculating the Hessian to precondition the gradient vector before using it for weight update.} Unfortunately in the context of zeroth-order optimization, the gradient itself is estimated, so we cannot directly compute the Hessian.

%However, loss functions of LLMs often exhibit heterogeneous curvatures across different parameter dimensions, as documented in recent studies \citep{sagun2017eigenvalues, Hessian_Eigenvalue_Density, Adaptive_Methods_Good_for_Attention}. This significant difference in curvature can lead to instability or decelerated training. For instance, parameters with larger curvature may experience same update magnitudes as parameters with smaller curvature, thereby failing to converge quickly, as shown in Figure~\ref{fig:3D}. The Hessian can be leveraged in optimizer and effectively adjust the magnitude of the parameter updates \citep{Liu2023sophia:, ADAHESSIAN, anil2021scalable}, solving the above dilemma. Unfortunately, in the context of zeroth-order optimization, one cannot directly compute the Hessian atop first-order derivatives.

\begin{figure*}[h]
%\vskip -0.1in
\centering
% \scalebox{0.70}{
\includegraphics[width=1\linewidth]{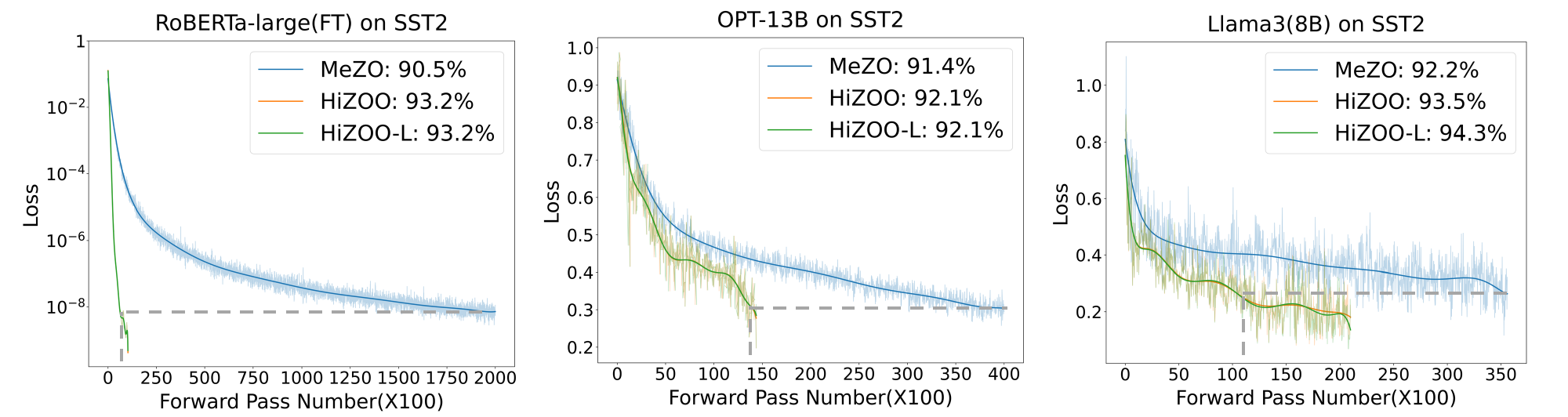}
\vspace{-7mm}
\caption{Performance of MeZO, HiZOO and HiZOO-L on SST2 task, when fine-tuning RoBERTa-large, OPT-13B, Llama3(8B) models. HiZOO can achieve 8$\times$ speedup and 1.55\% absolute accuracy improvement compared with MeZO.}
\label{fig:intro}
\vskip -0.02in
\end{figure*}

In light of above, we propose \name, as shown in Figure~\ref{fig:3D} (left), which estimates the diagonal Hessian by one more forward pass. HiZOO can act as a pre-conditioner, directly adjusting the update size of different parameters according to their curvatures. So that it can improve the model convergence when encountered with heterogeneous curvatures. As shown in Figure~\ref{fig:intro}, \nameo can significantly reduce number of training steps and improve model accuracy. Here we summarize our key contributions as follows:%Consequently, when facing heterogeneous curvatures, parameters with greater curvature will experience more substantial update magnitudes. 

% \begin{figure*}[h]
% %\vskip -0.1in
% \centering
% \scalebox{1}{
% \includegraphics[width=1\linewidth,trim=5 0 100 290,clip]{Figure/SST2.pdf}}
% \vspace{-5mm}
% \caption{\nameo achieves $8\times$ speedup and 1.55\% accuracy improvement over MeZO on average across different models. All the comparative experiments are conducted with the same parameters shown in Appendix~\ref{appendix_robert}.} 
% \label{fig:intro}

% \vskip -0.05in
% \end{figure*}

\vspace{-3mm}
\begin{enumerate}
    \item In this work, we estimate the Hessian in zeroth-order optimizer to fine-tune LLMs for the first time. Our HiZOO reduces the total number of forward passes required for model convergence and achieves better accuracy. By utilizing diagonal Hessian, HiZOO reduces the corresponding memory cost from $\mathcal{O}(d^2)$ to $\mathcal{O}(d)$. Furthermore, we propose HiZOO-L, reducing the memory usage of Hessian to \textbf{10\%} of the MeZO. %We also propose two variants to further decrease the memory cost and accelerate the model convergence. %What's more, \nameo utilizes diagonal Hessian and reduces the corresponding memory cost from $\mathcal{O}(d^2)$ to $\mathcal{O}(d)$.% 
    
    \item We provide theoretical analysis to prove that \nameo provides an unbiased estimation of the Hessian. Also, we illustrate how HiZOO utilizes Hessian to improve the convergence process by visualizing the optimization trajectories on test functions. % in even cases with heterogeneous curvatures across parameter dimensions.% (see Algorithm \ref{alg:MeZO_Hessian} in Section \ref{sec_methods} and Algorithm \ref{alg:MeZO_Hessian_multi} in Appendix \ref{app_MeZO_Hessian_variants}).
    
    \item We conduct extensive experiments across different models (RoBERTa-large, OPT, Llama3 and Phi-2) with scales from 350M to 66B, different methods (FT, LoRA, prefix), and different downstream tasks (classification, multiple-choice, and generation) to verify the effect of the \name. For example, on SST2 task HiZOO achieves a better accuracy and \textbf{8}\boldsymbol{$\times$} speedup over MeZO on average across different models. Even on OPT-66B, \nameo outperforms better than MeZO with up to \textbf{5.1\%} absolute improvement.% and 2\% improvement on average. %We also demonstrate \name ’s compatibility with full-parameter tuning, LoRA and prefix-tuning in Section~\ref{sec_experiments}. %Specifically, \nameo(prefix) outperforms MeZO (prefix) on most of the datasets and tasks across all model scales.
    %Especially on SST-2 dataset, \nameo can reduce the training time by up to 30 times (from 100k steps to 3k steps), achieving the same performance. What's more, \nameo finally improves the accuracy by 3\% on SST-2, even better than standard full-parameter fine-tuning.
    
    %\item 
    
    \item Further exploration in Section~\ref{section_F1} showcases that \nameo can achieve better performance in optimizing non-differentiable objectives such as F1 score. Specifically, \nameo significantly outperforms MeZO 's results with \textbf{6.5\%} absolute on average.% when fine-tuning RoBERTa-large and OPT.
\end{enumerate}

\section{Related Works}
\label{sec_related_works}

Here we present a concise overview on optimizers used in fine-tuning LLMs(details in Appendix \ref{app_related_works}).
%In light of space constraints, here we present a concise overview of related works on optimizers used in fine-tuning LLMs. For a more detailed review, please refer to Appendix \ref{app_related_works}.

%\subsection{First-Order Optimizer in LLMs}
\textbf{First-Order adaptive optimizer used in fine-tuning LLMs}~~~~Optimization methods have consistently been a popular research domain. Adaptive first-order optimizer, such as Gradient Descent (GD), Momentum, Adagrad \citep{Adagrad}, are fundamental in many areas like computer vision, natural languagle processing (NLP). Among them, Adam \citep{Adam} plays a dominant role due to its fast convergence and is often chosen for training and fine-tuning LLMs. AdamW \citep{adamw} improves upon Adam by adding the weight decay to alleviate overfitting. But both of them requires lots of memory cost due to the backpropagation process. This issue has become increasingly critical as the number of LLM parameters skyrockets.%To training neural networks with large batch, LAMB~\citep{LAMB} introduced a layer-wise adaptation strategy.% to accelerate training process. 

%\subsection{First-Order Optimizer Enhanced with Hessian}
\textbf{Enhanced optimizers with Hessian}~~~~On the other hand, researchers incorporated second-order information (Hessian) to provide richer guidance for gradient descent during the training. For example BROYDEN \citep{second-order-curvature-information1} ,  Nesterov \& Polyak~\citep{second-order-curvature-information2} and Conn et al. \citep{second-order-curvature-information3} utilized curvature information to pre-condition the gradient; Magoulas et al.~\citep{diagonal_Hessian1} applied diagonal Hessian as the pre-conditioner; 
 Martens \citep{conjugate_gradient} approximated the Hessian with conjugate gradient. Sophia \citep{Liu2023sophia:} used a light-weight estimate of the diagonal Hessian for pre-training LLMs. Despite their potential, above optimizers require the enormous GPU-memory cost. Additionally, these methods can only be used when first-order gradients are available.

%\subsection{Zeroth-Order Optimizer}
\textbf{Zeroth-Order Optimizer}~~~~Zeroth-order optimizers, with just forward passes to estimate the gradient, can greatly reduce the memory consumption. It appears in a wide range of applications where either the objective functions is implicit or its gradient is impossible or expensive to obtain. Methods like SPSA~\citep{ZO_GradientEstimator_SPSA} have been shown to perform well in non-convex multi-agent optimization~\citep{distributed_ZO1, distributed_ZO2} or generating black-box adversarial examples \citep{black_box_adversial_ZO1, black_box_adversial_ZO2, black_box_adversial_ZO3, black_box_adversial_ZO4}. Recently, MeZO \citep{mezo} first adapted the classical ZO-SGD method to fine-tune LLM, achieving comparable performance with significant memory reduction. Then \cite{zobenchmark} proposed a wider array of ZO optimization techniques. However, these methods often struggle with heterogeneous curvatures.

\section{Methods}
\label{sec_methods}

In the following, we briefly introduce the classical ZO gradient estimator SPSA~\citep{ZO_GradientEstimator_SPSA}, which is used in MeZO. Then we describe how HiZOO estimates diagonal Hessian and cooperates with ZOO. We also provide detailed proof for our method.

\subsection{Preliminaries}

\begin{definition} Simultaneous Perturbation Stochastic Approximation or SPSA

Given a model with parameters $\theta \in \mathbb{R}^d$ and loss function $\mathcal{L}$, SPSA estimates the gradient on a minibatch $\mathcal{B}$ , based on the concepts of sampling and differencing, as shown below:

{\setlength\abovedisplayskip{-0.3cm}
\setlength\belowdisplayskip{-0.2cm}
 \begin{align*}
     g^{\prime}_\mu(\theta_t) 
     =
     \frac{\mathcal{L}(\theta_t + \mu u; \mathcal{B}) - \mathcal{L}(\theta_t - \mu u; \mathcal{B})}{2\mu} u 
        \approx& u u^\top \nabla \mathcal{L}(\theta_t; \mathcal{B}),
 \end{align*} 
		\label{eq:spsa}}
  
where $u \in \mathbb{R}^d$ and is sampled from $\mathcal{N}(0,I_d)$, $\mu$ is the \emph{perturbation scale}. The $n$-SPSA gradient estimate averages $g_\mu(\theta)$ over $n$ randomly sampled $u$. 
	\label{def:spsa}
\end{definition}

\subsection{Hessian Informed Zeroth-Order Optimization}
%\subsection{Natural Gradient Estimation}

% \begin{figure}[t] % {1.0\textwidth}
% \centering
\begin{algorithm}[t]
\begin{algorithmic}[1]
%\setstretch{1.05}
\Require parameters $\theta \in \mathbb{R}^d$, loss $L : \mathbb{R}^d \rightarrow \mathbb{R}$, step budget $T$, perturbation scale $\mu$, learning rate schedule ${\eta_t}$, smooth scale $\alpha_t$, diagonal Hessian $\Sigma_{0}$

  %\vspace{0.1cm}
\For{$t=1,...,T$}
\State Sample batch $ \mathcal{B} \subset \mathcal{D}$ and random seed $s$
    \State $\ell \leftarrow \mathcal{L}(\theta; \mathcal{B})$ 
    \State $\theta$ $\leftarrow$ PerturbParameters($\theta$, $\mu$,            $\Sigma^{1/2}_{t-1}$, $s$)
    \State $\ell_{+} \leftarrow \mathcal{L}(\theta; \mathcal{B})$
    \State $\theta$ $\leftarrow$ PerturbParameters($\theta$, $-2 \mu$, $\Sigma^{1/2}_{t-1}$, $s$)
    \State $\ell_{-} \leftarrow \mathcal{L}(\theta; \mathcal{B})$
    %\vspace{1mm}\STATE \CommentSty{//Reset parameters before descent}\\
    \State $\theta$ $\leftarrow$ PerturbParameters($\theta$, $\mu$, $\Sigma^{1/2}_{t-1}$, $s$)
    \Comment{Reset parameters before descent}
    %\vspace{1mm}
    \State $\Sigma^{\prime}_{t} = \frac{1}{2 \mu ^2 } (\ell_{+} + \ell_{-} - 2 \ell )(\Sigma^{-1/2}_{t-1} u_i u_i ^{\top} \Sigma^{-1/2}_{t-1})$
    \Comment{Update diagonal Hessian}% by Eq.~\eqref{eq:Hessian_update}}\\
    %\vspace{1mm}\STATE \CommentSty{//Update diagonal Hessian}\\
    %\vspace{1mm}
    \State $\Sigma_{t}^{-1} = (1-\alpha_t)\Sigma_{t-1}^{-1} + \alpha_t  \left| diag(\Sigma^{\prime}_{t})\right|$   
    \State projected\_grad $\leftarrow (\ell_{+} - \ell_{-}) \ast \Sigma^{1/2}_t /2\mu$
    
    %\STATE \CommentSty{//For sampling $u_i$}\\
    \State Reset random number generator with seed $s$ 
    \Comment{For sampling $u_i$}
    %\STATE  \CommentSty{//Update parameters in place}\\
    \For{$\theta_i \in \theta$}
        \State Sample $u_i \sim \mathcal{N}(0,I_d)$
        \State $\theta_i \leftarrow \theta_i - \eta_t \ast$  projected\_grad $\ast u_i$
    \EndFor
\EndFor
%\Statex
\Function{PerturbParameter}{$\theta$, $\mu$, $\Sigma^{1/2}_t$, $s$}
    %\STATE \CommentSty{//For sampling $u_i$}\\
    \State Reset random number generator with seed $s$  
    \Comment{For sampling $u_i$}
    %\STATE  \CommentSty{//Modify parameters in place}\\
    \For{$\theta_i \in \theta$}
        \State Sample $u_i \sim \mathcal{N}(0,I_d)$
        \State $\theta_i \leftarrow \theta_i + \mu \Sigma^{1/2}_t u_i$
        \Comment{Modify parameters in place}
    \EndFor
    \State \Return $\theta$
\EndFunction
  \caption{\name}
  \label{alg:MeZO_Hessian}
  \end{algorithmic}
\end{algorithm}
% \vspace{-25pt}
% \end{figure}

We will present how to estimate Hessian inverse matrix $\Sigma$ in detail in Section~\ref{Hessian_Estimator}. Given $\Sigma$, then we can construct the following descent direction:

{\setlength\abovedisplayskip{-0.3cm}
\setlength\belowdisplayskip{-0.2cm}
\begin{small}
\begin{equation}
\begin{split}
g_{\mu}(\theta_t) = \sum\limits_{i=1}^{n} \frac {\mathcal{L}(\theta_t+\mu \Sigma^{1/2}_t u_i; \mathcal{B}) - \mathcal{L}(\theta_t-\mu \Sigma^{1/2}_t u_i; \mathcal{B}) }{2\mu \cdot n} \cdot \Sigma_t^{1/2} u_i.
\end{split}
 \label{eq:gradient_estimate}
\end{equation}
\end{small}
}

With the above descent direction, we can update $\theta_t$ as follows:

{\setlength\abovedisplayskip{-0.5cm}
\setlength\belowdisplayskip{-0.2cm}
\begin{equation}
\theta_{t+1} = \theta_t -\eta_t g_{\mu}(\theta_t).
 \label{eq:parameter_update}
\end{equation}}

It's guaranteed that $g_\mu(\theta)$ can estimate the descent direction by the following equation:

{\setlength\abovedisplayskip{-0.6cm}
\setlength\belowdisplayskip{-0.05cm}
\begin{align}
    \mathbb{E} \left[\mathcal{L}(\theta_{t+1}; \mathcal{B})\right] 
    &=
    \mathcal{L}(\theta_t; \mathcal{B}) - \eta_t \mathbb{E}\left[\langle \nabla \mathcal{L}(\theta_t; \mathcal{B}), g_\mu(\theta_t)\rangle\right] + \mathcal{O}(\eta_t^2) \nonumber \\
    &=
    \mathcal{L}(\theta_t; \mathcal{B}) - \eta_t \frac{1}{b} \mathbb{E}\left[\sum_{i=1}^b\langle \nabla \mathcal{L}(\theta_t; \mathcal{B}), \Sigma_t^{1/2}u_iu_i^\top \Sigma_t^{1/2} \nabla \mathcal{L}(\theta_t; \mathcal{B})\rangle\right] \nonumber
    + \mathcal{O}(\eta_t^2) + \mathcal{O}(\mu) \nonumber\\
    &=
    \mathcal{L}(\theta_t; \mathcal{B}) - \eta_t \|\Sigma_t^{1/2}\nabla\mathcal{L} (\theta_t; \mathcal{B})\|^2 +  \mathcal{O}(\eta_t^2) + \mathcal{\mu} \nonumber,
\end{align}}

\begin{comment}
\begin{equation}
\begin{aligned}
    &\mathbb{E} \left[\mathcal{L}(\theta_{t+1})\right] \\
    =& 
    \mathcal{L}(\theta_t) - \eta_t \mathbb{E}\left[\langle \nabla \mathcal{L}(\theta_t), g_\mu(\theta_t)\right] + \mathcal{O}(\eta_t^2) \\
    =&
    \mathcal{L}(\theta_t) - \eta_t \frac{1}{b} \mathbb{E}\left[\sum_{i=1}^b\langle \nabla \mathcal{L}(\theta_t), \Sigma_t^{1/2}u_iu_i^\top \Sigma_t^{1/2} \nabla \mathcal{L}(\theta_t)\right] \\
    &+ \mathcal{O}(\eta_t^2) + \mathcal{O}(\mu)\\
    =&
    \mathcal{L}(\theta_t) - \eta_t \|\Sigma_t^{1/2}\nabla (\theta_t)\|^2 +  \mathcal{O}(\eta_t^2) + \mathcal{\mu}.
\end{aligned}
\end{equation}
\end{comment}

where the first and second equality are both from the Taylor's expansion.
Above equation shows that when $\eta_t$ is properly chosen, $g_\mu(\theta)$ can accurately estimate the direction of gradient descent, which is the key to the success of fine-tuning large language models.

\vspace{-2pt}
\subsection{Diagonal Hessian Estimator}
\label{Hessian_Estimator}
\vspace{-2pt}
%We uses a diagonal Hessian-based pre-conditioner, which
%directly adjusts the update size of different parameter dimensions according to their curvatures. Similar to the gradient of the mini-batch loss function, the estimated diagonal Hessian can also have large noise. Inspired by the EMA of moments of gradients in Adam, we also denoise the diagonal Hessian estimates with EMA across iterations. We update the EMA every k steps, resulting in the following update rule for the diagonal Hessian estimate:

Given a model with parameters $\theta \in \mathbb{R}^d$, storing the exact full spectral Hessian ($d \times d$) requires $\mathcal{O}(d^2)$ memory~\citep{full_hessian1, full_hessian2, full_hessian3}, which is sufficient but never necessary. In HiZOO, we just estimate and retain only the diagonal Hessian which requires  $\mathcal{O}(d)$ memory. It serves as a pre-conditioner to scale the direction and magnitude of the model parameter updates according to their respective curvatures. 

Drawing from the lemma presented in MiNES \citep{MiNES}:

{\setlength\abovedisplayskip{-5cm}
\setlength\belowdisplayskip{-5cm}
\begin{equation}
 \frac{1}{2} \cdot \mathbb{E}_u (u^{\top} \Sigma^{1/2} H \Sigma^{1/2} u \cdot (\Sigma^{-1/2}u u^{\top} \Sigma^{-1/2} - \Sigma^{-1}))=H,
 \label{eq:MiNES}
\end{equation}}

where $H$ is the Hessian $\nabla^2 \mathcal{L}(\theta; \mathcal{B})$ and $\Sigma$ is a positive definite matrix.% with $\Sigma^{-1} \approx \nabla ^2 \mathcal{L}(\theta; \mathcal{B})$.

Thus, we can approximate the diagonal Hessian by the zeroth order oracles.
Firstly, we will access to the $\mathcal{L}(\theta + \mu \Sigma^{1/2} u; \mathcal{B})$, $\mathcal{L}(\theta -\mu \Sigma^{1/2} u; \mathcal{B})$ and $\mathcal{L}(\theta; \mathcal{B})$.
Through the Taylor's expansion, we yield the following results:

{\setlength\abovedisplayskip{-0.3cm}
\setlength\belowdisplayskip{-0.1cm}
% \begin{equation}\label{eq:aa}
% \begin{aligned}
\begin{align*}
    \mathcal{L}(\theta + \mu \Sigma^{1/2} u; \mathcal{B}) =
    \mathcal{L}(\theta; \mathcal{B}) + \mu \langle \mathcal{L}(\theta; \mathcal{B}), \Sigma^{1/2} u\rangle 
    + \frac{\mu^2}{2} u^\top \Sigma^{1/2}\nabla^2 \mathcal{L}(\theta; \mathcal{B}) \Sigma^{1/2} u + \alpha(\theta, \mu \Sigma^{1/2} u).
% \end{aligned}    
% \end{equation}}
\end{align*}}

Similarly, we also have:
{\setlength\abovedisplayskip{0cm}
\setlength\belowdisplayskip{-0.1cm}
\begin{align*}
   \mathcal{L}(\theta - \mu \Sigma^{1/2} u; \mathcal{B}) 
    = 
    \mathcal{L}(\theta; \mathcal{B}) - \mu \langle \mathcal{L}(\theta; \mathcal{B}), \Sigma^{1/2}u\rangle 
    + \frac{\mu^2}{2} u^\top \Sigma^{1/2}\nabla^2 \mathcal{L}(\theta; \mathcal{B}) \Sigma^{1/2}u + \alpha(\theta, -\mu\Sigma^{1/2} u).
\end{align*}}
%where $\alpha(\theta, -\mu\Sigma^{1/2} u)$ is of order $\mathcal{O}(\mu^3)$.

%Make $\Delta \mathcal{L} = \mathcal{L}(\theta + \mu \Sigma^{1/2} u) + \mathcal{L}(\theta - \mu\Sigma^{1/2} u) - 2\mathcal{L}(\theta)$,

Then we can calculate the difference $\Delta \mathcal{L} $ by:

\begin{comment}
\begin{align*}
 \Delta \mathcal{L} = &2\mathcal{L}(\theta) + \mu^2 u^\top \Sigma^{1/2} \nabla^2 \mathcal{L}(\theta) \Sigma^{1/2} u \\
 &+ \alpha(\theta, \mu \Sigma^{1/2}u) + \alpha(\theta, -\mu\Sigma^{1/2} u) -2\mathcal{L}(\theta)\\
 =&
 \mu^2 u^\top \Sigma^{1/2} \nabla^2 \mathcal{L}(\theta) \Sigma^{1/2} u + \alpha(\theta, \mu \Sigma^{1/2}u)+ \alpha(\theta, -\mu\Sigma^{1/2} u).
\end{align*}
\end{comment}

{\setlength\abovedisplayskip{-0.3cm}
\setlength\belowdisplayskip{0cm}
\begin{align*}
 \Delta \mathcal{L} &= \mathcal{L}(\theta + \mu \Sigma^{1/2} u; \mathcal{B}) + \mathcal{L}(\theta - \mu\Sigma^{1/2} u; \mathcal{B}) - 2\mathcal{L}(\theta; \mathcal{B})\\
 &=
 \mu^2 u^\top \Sigma^{1/2} \nabla^2 \mathcal{L}(\theta; \mathcal{B}) \Sigma^{1/2} u + \alpha(\theta, \mu \Sigma^{1/2}u) + \alpha(\theta, -\mu\Sigma^{1/2} u).
\end{align*}}

%Let $\Delta \mathcal{L} = \mathcal{L}(\theta + \mu \Sigma^{1/2} u) + \mathcal{L}(\theta - \mu\Sigma^{1/2} u) - 2\mathcal{L}(\theta)$.

Since $\alpha(\theta, \mu \Sigma^{1/2}u)$ and $\alpha(\theta, -\mu\Sigma^{1/2} u)$ are of order $\mathcal{O}(\mu^3)$, we can obtain that:

{\setlength\abovedisplayskip{-0.3cm}
\setlength\belowdisplayskip{-0.15cm}
\begin{align*}
    \frac{ \Delta \mathcal{L}}{\mu^2} 
    =
   u^\top \Sigma^{1/2} \nabla^2 \mathcal{L}(\theta; \mathcal{B}) \Sigma^{1/2} u + \mathcal{O}(\mu).
\end{align*}}

\begin{comment}
\begin{align*}
    &\frac{ \mathcal{L}(\theta + \mu \Sigma^{1/2} u) + \mathcal{L}(\theta - \mu\Sigma^{1/2} u) - 2\mathcal{L}(\theta)}{\mu^2} \\
    =&
   u^\top \Sigma^{1/2} \nabla^2 \mathcal{L}(\theta) \Sigma^{1/2} u + \mathcal{O}(\mu).
\end{align*}
\end{comment}

Upon substituting the above results into the left side of the Eq.~\eqref{eq:MiNES}, we arrive at:

{\setlength\abovedisplayskip{-0.3cm}
\setlength\belowdisplayskip{0cm}
\begin{align*}
\frac{1}{2} \mathbb{E} \bigg[ \frac{\Delta \mathcal{L}}{\mu^2} \cdot 
\left( \Sigma^{-1/2}u u^{\top} \Sigma^{-1/2} - \Sigma^{-1} \right) \bigg] = \nabla^2 \mathcal{L}(\theta; \mathcal{B}) + \mathcal{O}(\mu).
\end{align*}}

\begin{comment}
\begin{align*}
&\frac{1}{2} \mathbb{E} \bigg[ \frac{\mathcal{L}(\theta + \mu \Sigma^{1/2} u) + \mathcal{L}(\theta - \mu\Sigma^{1/2} u) - 2\mathcal{L}(\theta)}{\mu^2} \cdot \\
&\left( \Sigma^{-1/2}u u^{\top} \Sigma^{-1/2} - \Sigma^{-1} \right) \bigg] = \nabla^2 \mathcal{L}(\theta) + \mathcal{O}(\mu).
\end{align*}
\end{comment}

Therefore, by generalizing above equation to the multi-sampling version, we can approximate the diagonal Hessian $\nabla^2 \mathcal{L}(\theta)$ at $\theta$ by:

{\setlength\abovedisplayskip{-1cm}
\setlength\belowdisplayskip{-0.4cm}
\begin{equation}
\begin{aligned}
 \Sigma^{\prime}_t(\theta) =
\frac{1}{2n } \sum\limits_{i=1}^{n} \bigg[ \frac{\Delta \mathcal{L}} {\mu ^2}
\cdot \bigg(\Sigma^{-1/2}_t u_i u^{\top}_i \Sigma^{-1/2}_t - \Sigma^{-1} \bigg) \bigg],
\end{aligned}
 \label{eq:Hessian_update}
\end{equation}}

\begin{comment}
\begin{equation}
\begin{aligned}
& \Sigma^{\prime}_t(\theta) =\\
&\frac{1}{2 n \mu ^2 } \sum\limits_{i=1}^{n} \bigg[\bigg(\mathcal{L}(\theta+\mu \Sigma^{1/2}_t u_i) + \mathcal{L}(\theta-\mu \Sigma^{1/2}_t u_i)  -2\mathcal{L}(\theta)\bigg)\\
&\cdot \Sigma^{-1/2}_t u_i u^{\top}_i \Sigma^{-1/2}_t \bigg].
\end{aligned}
 \label{eq:Hessian_update}
\end{equation}
\end{comment}

where $n$ denotes the number of sampling instances for $u$, indicating the frequency of estimation per step. A larger $n$ diminishes the variance of the diagonal Hessian estimation and simultaneously increases computational overhead. Here we adopt $n=1$ as the default setting and present the pseudo-code of \nameo in Algorithm \ref{alg:MeZO_Hessian}. Further experimental investigation into the impact of varying $n$ is available in the Section \ref{Hyperparameter Analysis}.

Above equation shows that we can approximate the diagonal entries of $\nabla^2\mathcal{L}(\theta; \mathcal{B})$ by $\mathrm{diag}(\Sigma_t'(\theta))$, requiring just one more forward pass per step compared with MeZO.

Due to the presence of noise in the calculation of the Hessian, we utilize exponential moving average (EMA) to denoise the diagonal Hessian estimation.%, which commonly used in moments of gradients of Adam.

{\setlength\abovedisplayskip{-0.5cm}
\setlength\belowdisplayskip{-0.3cm}
\begin{equation}
\Sigma_{t+1}^{-1} = (1-\alpha_t)\Sigma_{t}^{-1} + \alpha_t  \left| \mathrm{diag}(\Sigma^{\prime}_{t})\right|.
 \label{eq:Hessian_EMA}
\end{equation}}

In the above equation, we firstly initial the $\Sigma_0 = I_d$ and update it every step with $\mathcal{O}(d)$ memory cost all the time. We also use $| \mathrm{diag}(\Sigma^{\prime}_{t})|$ to keep all entries of $\Sigma_t$ to be non-negative.

To further reduce Hessian memory consumption, we propose HiZOO-L to maintain it in a low-rank subspace, motivated by Adafactor~\citep{adafactor}. 
For $\hat{\Sigma}^{-1} \in \mathbb{R}^{p \times q}$, we will store two low-rank matrices $R \in \mathbb{R}^{p \times k} $ and $C \in \mathbb{R}^{k \times q} $ with $k=1$. %such that $\hat{\Sigma}^{-1} \approx R C$ in each step. 
Specifically, we can get $\hat{\Sigma}^{-1}$ by:

{\setlength\abovedisplayskip{-0.3cm}
\setlength\belowdisplayskip{0cm}
\begin{align*}
 \hat{\Sigma}^{-1}_{t} = (R_{t} * C_{t})/(1_{p}^{\top}*R_{t}),
\end{align*}}

where $1_{p}=(1,\cdots,1)\in \mathbb{R}^{p}$ denotes a  column vector of $p$ ones. Then in each step, we will update the $R$ and $C$  separately:

{\setlength\abovedisplayskip{-0.4cm}
\setlength\belowdisplayskip{0cm}
\begin{align*}
 R_t^{-1} = (1-\alpha_t)R_{t-1}^{-1} + \alpha_t \left| diag(\hat{\Sigma}^{\prime}_{t})\right| * 1_q,
\end{align*}}

{\setlength\abovedisplayskip{-0.2cm}
\setlength\belowdisplayskip{-0.2cm}
\begin{align*} 
 C_t^{-1} = (1-\alpha_t)C_{t-1}^{-1} + \alpha_t 1^{\top}_p * \left|   diag(\hat{\Sigma}^{\prime}_{t})\right|.
\end{align*}}

Detailed Algorithm can be seen in Appendix~\ref{app_MeZO_Hessian_variants}.

\subsection{Convergence Analysis}
\label{Convergence_Analysis}

%In Section~\ref{sec_methods}, we have briefly discussed that $g_\mu(x)$ can estimate the gradient descent direction. 
In this section, we will analyse the convergence based on the assumption of non-convex optimization (details in Appendix~\ref{app_convergence_proof}).

\begin{theorem}
Let the descent direction $g_\mu(\theta_t)$ defined as:

{\setlength\abovedisplayskip{-0.3cm}
\setlength\belowdisplayskip{-0.3cm}
\begin{small}
\begin{equation}\label{eq:gmu}
\begin{aligned}
    &g_\mu(\theta_t)
        =
        \sum_{i=1}^b \frac{\mL(\theta_t + \mu \Sigma_t^{1/2} u_i;\mathcal{B}_t) - \mL(\theta_t - \mu \Sigma_t^{1/2} u_i;\mathcal{B}_t)}{2b\mu} \Sigma_t^{1/2}u_i.
\end{aligned}
\end{equation}
\end{small}}

Based on Assumption~\ref{ass:L}-\ref{ass:beta}, if the update rule for $\theta$ is $\theta_{t+1} = \theta_t - \eta g_\mu(\theta_t)$ for a single step, then it's established that:
{\setlength\abovedisplayskip{0cm}
\setlength\belowdisplayskip{0cm}
\begin{equation}\label{eq:theta}
\begin{aligned}
    \EE\left[\mL(\theta_{t+1}) \mid \right] 
    \leq \mL(\theta_t) - \frac{\eta_t}{4} \|\nabla \mL(\theta_t)\|_{\Sigma_t}^2  
    + 2\eta_t^2L\left( \mathrm{tr}(\Sigma_t) +\beta_u \right)\sigma^2 + \cO(\mu^2).
\end{aligned}
\end{equation}}

Furthermore, given iteration number $T$, we choose the step size $\eta = \frac{1}{8\sqrt{T}L(\max_t\mathrm{tr}(\Sigma_t) +\beta_u)}$ and take $\theta_{\mbox{out}} = \theta_j$ with $j$ uniformly sampled from $\{1, \dots, T\}$.
Then, we have

\begin{small}
\begin{equation}\label{eq:Lout}
\begin{split}
\EE\left[\|\nabla \mL(\theta_{\mbox{out}})\|^2\right]
\leq
\frac{32L\left( \max_t \{\mathrm{tr}(\Sigma_t)\} +\beta_u \right) (\mL(\theta_1) - \mL(\theta_*))}{\sqrt{T}\beta_\ell } 
+ \frac{\sigma^2}{T^{3/2} \beta_\ell} + \cO\left(\mu^2\right),
\end{split}
\end{equation}
\end{small}

where $\mL(\theta_* )$ minimizes the function $\mL(\theta;)$. The above equation shows that as $T\to \infty$, \nameo can converge to the stationary point.
\end{theorem}

\begin{proof}
\label{proof}
Detailed proof can be found in Appendix~\ref{app_convergence_proof}.
\end{proof}

\subsection{Visualization of HiZOO on Test Functions}
\label{2D_curvature}

Despite above theoretical guarantee, \textbf{we still want to illustrate how HiZOO utilizes Hessian to improve the convergence process}. But it's impractical for large models to visualize their optimization trajectories. Therefore we choose three test functions (see details in Appendix~\ref{app_test_functions}) with heterogeneous curvatures across different parameters and visualize the optimization trajectories on them.% This visualization helps to demonstrate how HiZOO handles different curvature conditions compared to Adam and MeZO.

% \begin{itemize}
% \setlength{\itemsep}{1pt}
% \vspace{-3mm}
% \item Function (a)\footnote{Function (a) is from \citep{Liu2023sophia:}.}: $f(x,y) = 8(x-1)^2(1.3x^2+2x+1)+0.5(y-4)^2$
% \item Function (b): $f(x,y) = \lvert x \rvert + \lvert y \rvert$
% \item Function (c): $f(x,y) = 10000 x^2 + y^2$
% \vspace{-3mm}
% \end{itemize}

\begin{figure*}[h]
\vskip -0.1in
\centering
%\scalebox{0.90}{
\includegraphics[width=1\linewidth,trim=0 0 50 270,clip]{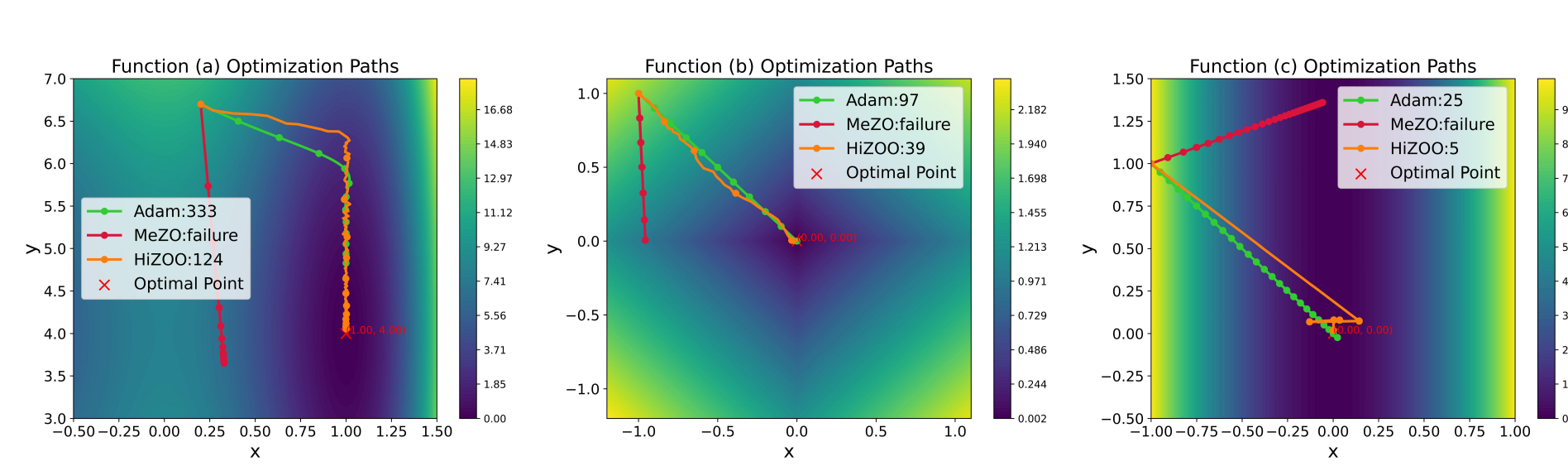}%}
\vspace{-8mm}
\caption{Optimization trajectories of Adam, MeZO and \nameo on 3 test functions. We have labeled the number of iterations required for the loss to drop to 0.1.}% In all cases, \nameo performs comparably to the Adam, but MeZO does not reach optimal points.}
\label{fig:2D}
\vskip -0.1in
\end{figure*}

As illustrated in Figure~\ref{fig:2D}, HiZOO and Adam both achieve better convergence on three functions, and HiZOO even requires less steps for convergence than Adam. However, MeZO only achieves effective convergence in either the $x$ or $y$ dimension, but not both, indicating a limitation in capturing this curvature difference. Particularly in function (c) curvature of $x$ is extremely bigger than $y$. In this case, HiZOO can sense this difference in parametric curvature and update the function along $x$ on purpose, achieving quicker convergence. In contrast, MeZO is very hard to converge.

\section{Experiments}
\label{sec_experiments}

Large language models are generally classified into two types: (1) Encoder-Decoder, also known as masked language models, such as BERT~\citep{BERT} and ALBERT~\citep{ALBERT}; (2) Decoder-Only, also recognized as generative language models, such as GPT family~\citep{gpt2, GPT3}, OPT~\citep{opt}, LLaMA~\citep{Touvron2023LLaMAOA}, Phi~\citep{li2023textbooks,gunasekar2023textbooks}. 
%Large language models are generally classified into two types: (1)Encoder-Decoder, also known as masked language models and representative examples are XLM\citep{XLM}, BERT\citep{BERT}, ALBERT\citep{ALBERT}, Roberta\citep{roberta} and so on; (2) Decoder-Only, also recognized as generative language models, including GPT family\citep{GPT3}, OPT\citep{opt}, LLaMA\citep{Touvron2023LLaMAOA}. 

%\input{Table/opt-13b}

To rigorously assess the universality and robustness of our \name, we have chosen models from each category for empirical testing. Additionally, we investigate FT and PEFT (LoRA~\citep{hu2022lora} and prefix~\citep{li-liang-2021-prefix}). Detailed experiment settings are presented in Appendix~\ref{appendix_robert}.

%We perform exhaustive experiments on medium-sized masked language models (e.g., RoBERTa-large, with 350 million parameters \citep{roberta}) and large autoregressive language models (e.g., OPT-13B, 30B, 66B \citep{opt}), employing few-shot and many-shot scenarios with prompts. Additionally, we investigate full-parameter tuning and Prompt-based Efficient Fine-Tuning (PEFT), which encompasses methods such as Low-Rank Adaptation (LoRA) \citep{hu2022lora} and prefix-tuning \citep{li-liang-2021-prefix}. Detailed methodologies are presented in Appendix. 

\subsection{Masked Language Models}
Firstly, we conduct experiments on RoBERTa-large 350M~\citep{roberta} on three NLP task paradigms: sentence classification, multiple choice and text generation. We follow the experimental setting~\citep{mezo} in studying the few-shot and many-shot, sampling $k$ examples per class for $k = 16$ (results in Table~\ref{tab:roberta_k16}) and $k = 512$ (results in Appendix~\ref{appendix_robert}). We did not utilize HiZOO-L here due to model's smaller parameter count.%which can reach the following observations and summaries.

\begin{table*}[t]
\vspace*{-2mm} 
\centering
\caption{Experiments on RoBERTa-large (350M parameters, k=16). PEFT represents using LoRA and prefix and we report the best result of them. All reported numbers are averaged accuracy (standard deviation) across 5 runs.}
\vspace*{2mm} 
\scalebox{0.85}{
    \begin{tabular}{lccccccc}
    \toprule
    Task Type & \multicolumn{1}{c}{\textbf{SST-2}} & \multicolumn{1}{c}{\textbf{SST-5}} & \multicolumn{1}{c}{\textbf{SNLI}}  & \multicolumn{1}{c}{\textbf{MNLI}} & \multicolumn{1}{c}{\textbf{RTE}} & \multicolumn{1}{c}{\textbf{TREC}} & \multicolumn{1}{c}{\textbf{Average}}\\
    & \multicolumn{2}{c}{------ sentiment ------} & \multicolumn{3}{c}{------ natural language inference ------} & \multicolumn{1}{c}{--- topic ---}\\
    \midrule
    Zero-shot & 79.0  & 35.5  & 50.2  & 48.8  & 51.4  & 32.0 & 49.5\\
        LP    &  76.0 ($\pm$2.8) & 40.3 ($\pm$1.9) & 66.0 ($\pm$2.7) & 56.5 ($\pm$2.5) & 59.4 ($\pm$5.3) & 51.3 ($\pm$5.5) & 58.3\\

\midrule
FT  & 91.9 ($\pm$1.8) & 47.5 ($\pm$1.9) & 77.5 ($\pm$2.6) & 70.0 ($\pm$2.3) & 66.4 ($\pm$7.2) & 85.0 ($\pm$2.5) & 74.9\\
PEFT & 91.9 ($\pm$1.0) & 47.7 ($\pm$1.1) & 77.2 ($\pm$1.3) & 67.7 ($\pm$1.4) & 66.6 ($\pm$2.0) & 85.7 ($\pm$1.3) & 72.8\\

\midrule
MeZO  & 90.5 ($\pm$1.2) & 45.5 ($\pm$2.0) & 68.5 ($\pm$3.9) & 58.7 ($\pm$2.5) & 64.0 ($\pm$3.3) & 76.9 ($\pm$2.7) & 67.4\\
MeZO (PEFT) & 91.4 ($\pm$0.9) & 45.8 ($\pm$2.0) & 71.6 ($\pm$2.5) & 62.1 ($\pm$2.5) & 61.0 ($\pm$3.9) & 80.3 ($\pm$3.6) & 68.7\\

\midrule
\name & \textbf{93.2} ($\pm$0.8) & 46.2 ($\pm$1.1) & \textbf{74.6} ($\pm$1.3) & \textbf{64.9} ($\pm$1.7) & \textbf{66.8} ($\pm$1.2) & 79.8 ($\pm$1.3) & \textbf{70.9}\\
\name (PEFT) & 92.3 ($\pm$1.2) & \textbf{47.2} ($\pm$1.1) & 71.1 ($\pm$1.1) & 62.1 ($\pm$1.7) & 65.4 ($\pm$1.2) & \textbf{82.0} ($\pm$2.0) & 70.0\\
    \bottomrule
    \end{tabular}}
    % With full-parameter fine-tuning, \nameo outperforms MeZO with \textbf{3.56\%} on average on all datasets.}
    \label{tab:roberta_k16}%
\end{table*}%

\begin{figure*}[ht]
%\vskip -0.1in
\centering
% \scalebox{0.70}{
\includegraphics[width=1\linewidth]{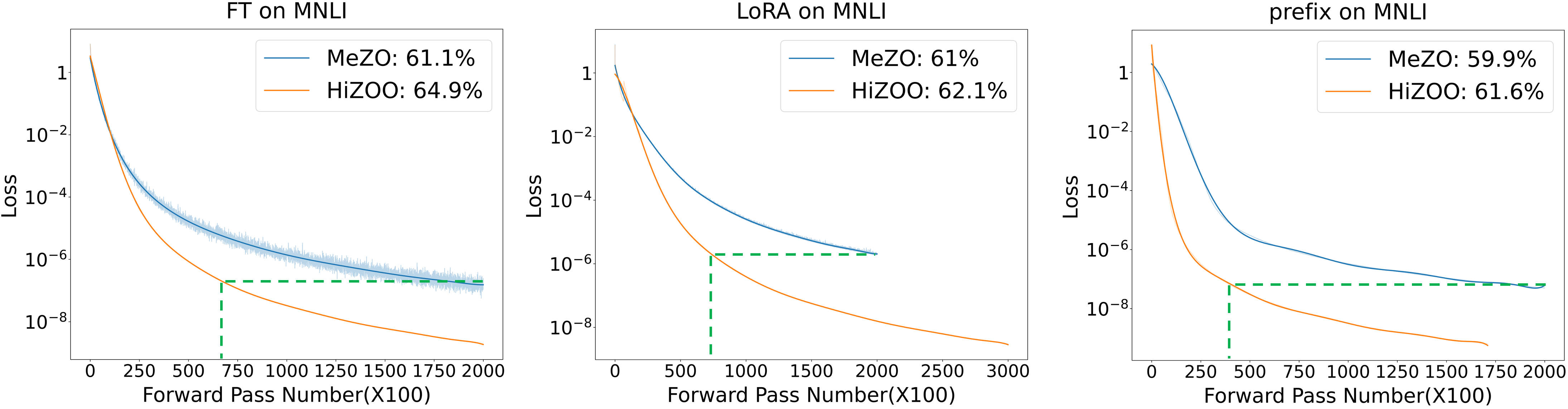}
\vspace{-6mm}
\caption{Training loss curves when using Adam, MeZO and HiZOO to fine-tune Roberta-large on MNLI. The evaluation accuracy curves can be found in Figure~\ref{fig:app_roberta_eval_acc} in Appendix~\ref{appendix_robert}.}
\label{fig:roberta-16}
\vskip -0.05in
\end{figure*}

% \begin{figure*}[h]
% \vskip -0.05in
% \centering
% % \scalebox{0.70}{
% \includegraphics[width=1\linewidth,trim=0 0 20 280,clip]{Figure/exp_MNLI.pdf}
% \vspace{-6mm}
% \caption{\nameo achieves a $4\times$ speedup on average over MeZO, both on FT and PEFT (prefix, LoRA) methods. All experiments are conducted using the same parameters in Appendix~\ref{appendix_robert}.}
% \label{fig:roberta-16}
% %\vskip -0.15in
% \end{figure*}

\textbf{\nameo greatly increases the convergence speed across full-parameter tuning, LoRA and prefix.} As shown in Figure \ref{fig:roberta-16}, \nameo achieves \textbf{4}\boldsymbol{$\times$} speedup over MeZO on average while getting the same training loss compared with MeZO. What's more, HiZOO finally achieves a \textbf{2.2\%} absolute accuracy improvement on MNLI better than MeZO. 
%More loss curves can be seen in Appendix~\ref{appendix_robert}.

\textbf{\nameo achieves better performance compared with MeZO.} Table~\ref{tab:roberta_k16} shows that \nameo outperforms MeZO's results with \textbf{3.5\%} absolute on average on all datasets across different tasks. Specifically, \nameo outperforms MeZO more than \textbf{6\%} in both the SNLI and MNLI dataset.  

% \begin{figure*}[h]
% %\vskip -0.1in
% \centering
% % \scalebox{0.70}{
% \includegraphics[width=1\linewidth]{Figure/OPT-13B.png}
% \vspace{-6mm}
% \caption{ OPT-13B results with zero-shot, in-context learning (ICL), MeZO, HiZOO and fine-tuning with Adam (FT). PEFT represents the best among LoRA and prefix. More details can be found in Appendix~\ref{app:opt}.}
% \label{fig:opt13b}
% \vskip -0.15in
% \end{figure*}

\begin{table*}[h]
\vspace{-5pt}
\caption{
        Experiments on three different models(with $1000$ examples). We highlight the best results between MeZO, HiZOO and HiZOO-L in bold to facilitate comparison.
    }
    \vspace{1.5pt}
\centering
% \setlength{\tabcolsep}{4pt}
% \resizebox{\textwidth}{!}{
\scalebox{0.9}{
    \begin{tabular}{lcccccccccccc}
    \toprule
    & \textbf{Model} & \textbf{Method}  & \textbf{SST-2}	& \textbf{RTE~} & \textbf{~CB~}  & \textbf{WSC~} & \textbf{WIC~}	 & \textbf{COPA}  & \textbf{MultiRC}& \textbf{Average}\\%& \textbf{DROP} \\
    %Task type & \multicolumn{7}{c}{------------------------ classification ------------------------} & \multicolumn{2}{c}{--- multiple choice ---} & generation\\ %\multicolumn{2}{c}{--- generation ---}\\
    \midrule

    %Adam(FT) {\fontsize{8}{9.6}\selectfont (12x memory)} & {92.0} & {70.8} &	{83.9} &	{77.1}	& {63.5} &	{70.1} &71.1 &{79.0} & {74.1} &{84.9}	\\%& {31.3} \\
    &Phi-2 &MeZO	&86.6	&67.1	&75.0	&59.6	&54.4		&86.0			&78.2 &72.4\\

    &Phi-2 &HiZOO     & 88.9    &69.0	&75.2	&62.5	&59.4	 &86.0   &79.2 &\textbf{74.3}\\

    &Phi-2 &HiZOO-L     &88.9    &68.9	 &75.2		&62.4	&59.2	 &86.0   &79.2 &74.2 \\
    \midrule
    
    &Llama3 &MeZO	&92.2	&74.4	&69.6		&63.5	&57.8		&88.0			&77.6&74.7\\

    &Llama3 &HiZOO     & 93.5    &75.1	&69.6		&63.5	&59.7	 &89.0  &78.2&\textbf{75.5}\\

    &Llama3 &HiZOO-L    &94.3    &75.1	 &69.6		&63.5	&57.7	 &89.0   &77.9 &75.3 \\
    \midrule
    %Adam(FT) {\fontsize{8}{9.6}\selectfont (12x memory)} & {92.0} & {70.8} &	{83.9} &	{77.1}	& {63.5} &	{70.1} &71.1 &{79.0} & {74.1} &{84.9}	\\%& {31.3} \\
    &OPT-13B &MeZO	&91.4	&66.1	&66.0	&63.5	&59.4		&88.0		&57.3 &70.2 \\

    &OPT-13B &HiZOO     & 92.1    &69.3	&69.6		&63.5	&59.4	 &89.0  &61.3&\textbf{72.1}\\

    &OPT-13B &HiZOO-L     &92.1 & 68.2  &67.9  &65.4 &59.4  &89.0  &61.1 &71.9\\

    \midrule
    %Adam(FT) {\fontsize{8}{9.6}\selectfont (12x memory)} & {92.0} & {70.8} &	{83.9} &	{77.1}	& {63.5} &	{70.1} &71.1 &{79.0} & {74.1} &{84.9}	\\%& {31.3} \\

    \bottomrule
    \end{tabular}}
\vspace{-5pt}
\label{tab:Llama3_phi2}
\end{table*}

\subsection{Auto-Regressive Language Models}
Then we extend experiments with Phi-2(2.7B), Llama3(8B) and OPT family on the same NLP task paradigms. The results of the experiment in Table~\ref{tab:Llama3_phi2} show that HiZOO outperforms MeZO in most cases. Also, we can see that HiZOO-L has only a slight decrease in accuracy. We also provide relative loss curves to show the better convergence process of our HiZOO in Appendix~\ref{app:Llama3_phi2_opt}. 

%The main results on OPT-13B are presented in Figure~\ref{fig:opt13b} and 
%Table~\ref{tab:opt-13b}. We can see that \nameo boosts MeZO's accuracy for 1.1\% accuracy on average using prefix-tuning. Specifically, HiZOO achieves 2.87\% and 3\% accuracy improvement on WSC and COPA tasks.

%Then we extend experiments with OPT family (13B, 30B, 66B) on the same NLP task paradigms. The main results on OPT-13B are presented in %Figure~\ref{fig:opt13b} and 
%Table~\ref{tab:opt-13b}. We can see that \nameo boosts MeZO's accuracy for 1.1\% accuracy on average using prefix-tuning. Specifically, HiZOO achieves 2.87\% and 3\% accuracy improvement on WSC and COPA tasks. %when fine-tuning the OPT-13B, \nameo(prefix) outperforms MeZO (prefix) on all datasets across classification and multiple choice tasks. 

%In addition, we utilize HiZOO to fine-tune the newly proposed model, Phi-2 (2.7B) and Llama3 (8B). Experiment results in Table~\ref{tab:Llama3_phi2} show that HiZOO outperforms MeZO in most cases. We also provide relative loss curves to show the better convergence process of our HiZOO in Appendix~\ref{app:Phi-2_Llama3}.% It's worth noting that since these two models are inherently very powerful, the improvement is relatively weak, with 0..

%\input{Table/opt66b}

\textbf{\nameo is capable of scaling to large models with up to 66B parameters, while preserving its exceptional performance.} As depicted in Table~\ref{tab_opt66}, on OPT-30B HiZOO outperforms MeZO with up to \textbf{2.9\%} increase and  \textbf{1.1\%} increase on average. Even scaling to OPT-66B, \name(prefix) still outperforms MeZO(prefix) with up to \textbf{5.1\%} increase and  \textbf{2.7\%} increase on average.% \textcolor{red}{Figure~\ref{fig:opt30} shows that at such scales \nameo can still effectively accelerate the process for convergence and reduce the training steps.}

\subsection{Training with Non-Differentiable Objectives}
\label{section_F1}
Our proposed \nameo employs gradient estimation to update parameters, allowing for the use of non-differentiable objectives for training. Following the setting of MeZO~\citep{mezo}, we conduct extensive experiments using F1 as optimization objective. The results presented in Table~\ref{tab_F1} indicate that our method outperforms MeZO by \textbf{6.54\%} absolute on F1 on average. %It's worth noting that on TREC the improvement is even more than 20\%.% and the original inability to converge using MeZO is solved. %What's more, our method reduces the convergence steps required for training for up to $1/10$ compared with MeZO, as shown in the Figure.

\begin{table}[htbp]
  \centering
  \vspace{-10pt}
  \begin{minipage}[t]{0.48\linewidth}

\centering
\caption{
    Experiments on OPT-30B (we use FT and prefix-tuning, report the best of them) and OPT-66B (we use prefix-tuning). 
}
\label{tab_opt66}
\vspace{1.8pt}
\resizebox{1.0\textwidth}{!}{ % 根据需要调整表格的实际大小
    \begin{tabular}{lcccccccc}
    \toprule
     Task  & \textbf{SST-2} & \textbf{RTE}  & \textbf{WSC} & \textbf{WIC}  & \textbf{Average} \\
    \midrule
    30B MeZO & 90.6 & 66.4  & \textbf{63.5} &59.1 & 69.9 \\
    30B HiZOO &\textbf{91.2} & \textbf{69.3} & \textbf{63.5} &\textbf{60.2} &\textbf{71.0}\\
    30B HiZOO-L &91.1 & 68.9  & \textbf{63.5} &59.8 &70.8\\

    \midrule
    66B MeZO & \textbf{93.6} & 66.4  & 57.7 &58.6 & 69.0 \\
    66B \name &\textbf{93.6} & \textbf{71.5}  & \textbf{60.6} &\textbf{61.1} &\textbf{71.7}\\
    66B HiZOO-L &\textbf{93.6} &71.0  &60.3 &60.9 &71.4\\

    \bottomrule
    \end{tabular}
}
  \end{minipage}
  \hfill
  \begin{minipage}[t]{0.48\linewidth}
    \centering
    \caption{
    Experiments on non-differentiable optimization objectives (F1). For classification ($k=512$), we use  full-parameter tuning and for SQuAD (1,000 examples), we use prefix tuning.
}
\label{tab_F1}
\vspace{1.8pt}
\resizebox{1.0\textwidth}{!}{ % 可以根据需要调整表格的实际大小
    \begin{tabular}{lccccc}
    \toprule
     Model & \multicolumn{4}{c}{RoBERTa-large (350M)} & OPT-13B \\
     \cmidrule{2-6}
     Task  & \textbf{SST-2} & \textbf{SST-5} & \textbf{SNLI} & \textbf{TREC}  & \textbf{SQuAD} \\
    \midrule
    Zero-shot & 79.0 & 35.5& 50.2 & 32.0  & 46.2\\
    MeZO & 92.7 &  48.9 & 82.7  & 68.6 & 78.5 \\
    \name & \textbf{94.9} &\textbf{52.9}   &\textbf{83.1}  & \textbf{90} &\textbf{83.21} \\
    \bottomrule
    \end{tabular}
}
  \end{minipage}
  \vspace{-10pt}
  
\end{table}

\subsection{Memory Usage and Time Efficiency Analysis}

\begin{figure*}[ht]
%\vskip -0.1in
\centering
% \scalebox{0.70}{
\includegraphics[width=1.1\linewidth]{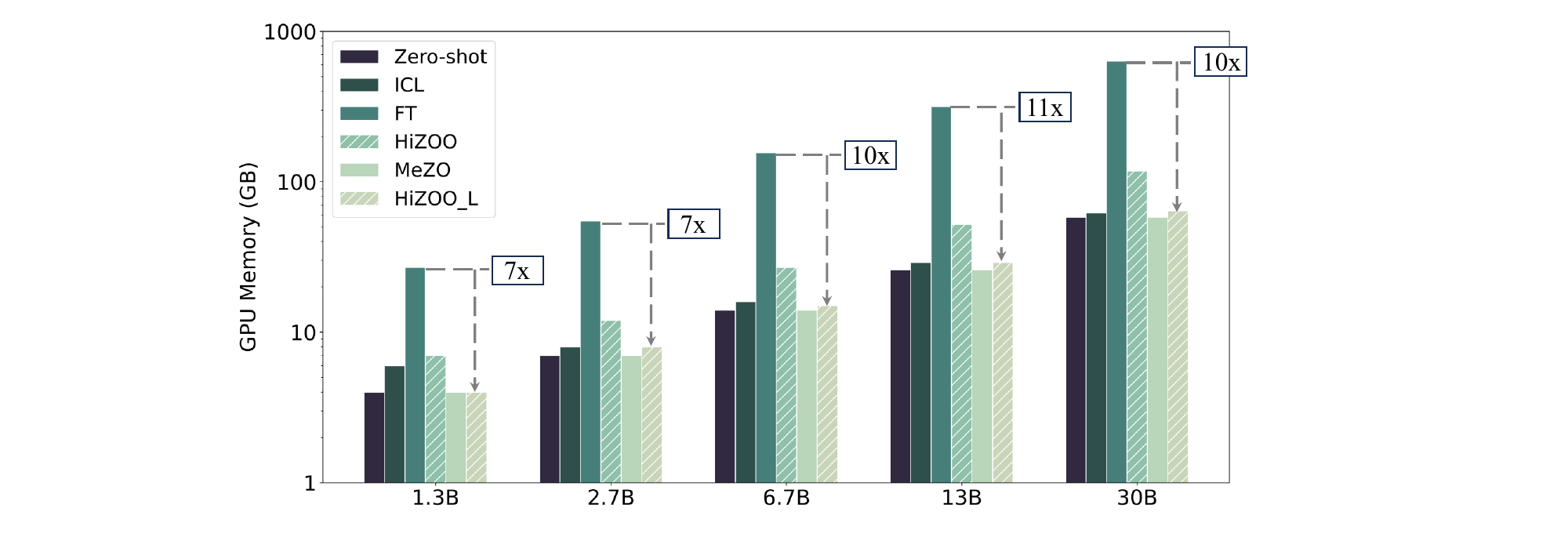}
\vspace{-8mm}
\caption{GPU memory consumption with different OPT models and tuning methods on MultiRC (400 tokens per example on average). More details can be found in Appendix~\ref{app:memory_time}.}
\label{fig:memory}
\vskip -0.05in
\end{figure*}

\textbf{Memory Usage}~~~~As shown in Figure~\ref{fig:memory}, \nameo increases the memory usage compared to MeZO because of the storage of the diagonal Hessian(refer to Appendix~\ref{app:memory_time} for detailed numbers). To further reduce memory consumption, we propose HiZOO-L, the low-rank implementation of HiZOO, motivated by Adafactor~\citep{adafactor}. Detailed Algorithm can be seen in Appendix~\ref{app_MeZO_Hessian_variants}. As a result, \name-L increases \boldsymbol{$\textless$} \textbf{10\%} memory more than MeZO, while maintaining the original performance of HiZOO. Specifically, using the same GPUs, \name-L allows for tuning a model that is 10 times larger than what is feasible with FT on average.
%, but it still holds great advantages over FT or FT (prefix) 

\textbf{Time Efficiency}~~~~We analyse the wall-clock time efficiencies and find that HiZOO and HiZOO-L spend $1.5\times$ time per step compared with MeZO, mainly from the extra forward pass, details in  Appendix~\ref{app:train_time}. However, HiZOO reduces total number of forward passes required for convergence. For example, HiZOO achieves a \textbf{8}\boldsymbol{$\times$} and \textbf{4}\boldsymbol{$\times$} speedup on SST2 and MNLI tasks.% What's more, HiZOO can yield a higher accuracy in many cases.

\subsection{Comparison with other ZO variants}
We also compare our HiZOO with a broader array of ZO optimization techniques~\cite{zobenchmark}. As shown in Table~\ref{benchmark}, our HiZOO outperforms all other ZO methods. Compared with ZO-Adam who leverages second-order moment to guide gradient descent, our HiZOO-L achieves a notable \textbf{4.3\%} absolute improvement, while using \textbf{50\%} of the GPU memory.

\begin{table*}[h]
\centering
\vspace{-15pt}
\caption{Performance comparison on SST2(Robert-Large and OPT-1.3B) and COPA(OPT-13B) using different ZO methods. Memory and runtime cost are multiples of
ZO-SGD.}% Comparison on memory and runtime can be seen in Appendix~\ref{app_benchmark_memory}.}
\vspace{7pt}
\scalebox{0.88}{
\begin{tabular}{lcccccccccc}
\toprule
\textbf{Model/Task} & \multicolumn{2}{c}{\textbf{Roberts-Large}} & \multicolumn{2}{c}{\textbf{OPT-1.3B}} & \multicolumn{2}{c}{\textbf{OPT-13B}} & \textbf{Average} &\textbf{Memory} & \textbf{Runtime}\\
\cmidrule(lr){2-3} \cmidrule(lr){4-5} \cmidrule(lr){6-7}
 & FT & prefix & FT & prefix & FT  & prefix &  \\
\midrule
%FO-SGD & 91.4 & 91.2 & 89.6 & 91.1 & 93.6 & 93.1 & 86.0 & 88.0 & 88.0 & 90.2 \\
%Forward-Grad & 90.1 & 89.7 & 89.5 & 90.3 & 90.3 & 90.0 & 87.0 & 89.0 & 79.0 & 88.3 \\
%\midrule
ZO-SGD & 89.4  & 90.0 & \textbf{90.8}  & \textbf{91.4} & \textbf{90.0}  & 79.0 & 88.4 & 1.0x & 1.0x\\
ZO-SGD-MMT & 89.6 & 89.1 & 85.2  & 91.2 & 87.0  & 85.0 & 87.8 & 1.56x & 1.0x\\
ZO-SGD-Cons & 89.6  & 89.1 & 88.3  & 88.1 & 82.0  & 84.0 & 86.8 & 1.0x & 2.49x\\
ZO-SGD-Sign & 52.5  & 53.6 & 87.2  & 89.5 & 80.0  & 78.0 & 73.4 & 1.0x & 1.0x \\
ZO-Adam & 89.8  & 90.2 & 84.4  & \textbf{91.4} & 82.0 &  79.0 & 86.1 & 2.47x & 1.04x \\
HiZOO & \textbf{93.2}  & \textbf{92.7} & 90.7  & \textbf{91.4} & 88.0  & \textbf{87.0} & \textbf{90.5} & 2.04x & 1.37x\\
HiZOO-L & 92.5  & \textbf{92.7} & 90.7  & \textbf{91.4} & 88.0  & \textbf{87.0} & 90.4 & 1.12x & 1.39x \\
% ZO-SGD & 89.4 & 90.8 & 90.0 & \textbf{90.8} & 91.4 & \textbf{91.4} & \textbf{90.0} & 87.0 & 79.0 & 88.9 & 1.0x & 1.0x\\
% ZO-SGD-MMT & 89.6 & 90.9 & 89.1 & 85.2 & 91.3 & 91.2 & 87.0 & 87.0 & 85.0 & 88.5 & 1.56x & 1.0x\\
% ZO-SGD-Cons & 89.6 & \textbf{91.6} & 89.1 & 88.3 & 91.8 & 88.1 & 82.0 & 88.0 & 84.0 & 88.1 & 1.0x & 2.49x\\
% ZO-SGD-Sign & 52.5 & 90.2 & 53.6 & 87.2 & 91.5 & 89.5 & 80.0 & \textbf{89.0} & 78.0 & 79.1 & 1.0x & 1.0x \\
% ZO-Adam & 89.8 & 89.5 & 90.2 & 84.4 & \textbf{92.3} & \textbf{91.4} & 82.0 & \textbf{89.0} & 79.0 & 87.5 & 2.47x & 1.04x \\
% HiZOO & \textbf{93.2} & 91.2 & \textbf{92.7} & 90.7 & 91.2 & \textbf{91.4} & 88.0 & 87.0 & \textbf{87.0} & \textbf{90.3} & 2.04x & 1.37x\\
% HiZOO-L & 92.5 & 91.1 & \textbf{92.7} & 90.7 & 91.2 & \textbf{91.4} & 88.0 & 87.0 & \textbf{87.0} & 90.2 & 1.12x & 1.39x \\
\bottomrule
\end{tabular}
}
\vspace{-7pt}
\label{benchmark}
\end{table*}

% \begin{table}[h]
% \centering
% \begin{tabular}{lcc}
% \toprule
% \textbf{Optimizer} & \textbf{Memory} & \textbf{Runtime Cost} \\
% \midrule
% ZO-SGD & 1.0x & 1.0x \\
% ZO-SGD-Cons & 1.0x & 2.49x \\
% ZO-SGD-Sign & 1.0x & 1.0x \\
% ZO-SGD-MMT & 1.56x & 1.0x \\
% Forward-Grad & 2.09x & 1.5x \\
% FO-SGD & 2.31x & 1.72x \\
% ZO-Adam & 2.47x & 1.04x \\
% FO-Adam & 3.83x & 1.75x \\
% HiZOO & 2.04x & 1.37x \\ 
% HiZOO-L & 1.12x & 1.39x \\
% \bottomrule
% \end{tabular}
% \caption{Comparison of Optimizer Memory and Runtime Cost as Multiples of ZO-SGD}
% \end{table}

% \begin{figure}[h]
% \vskip -0.1in
% \centering
% \scalebox{0.9}{
% \includegraphics[width=0.9\linewidth]{Figure/ablation_Hessian.png}}
% \vspace{-2mm}
% \caption{Experiments on influence of the value of Smooth scale $\alpha_t$ in Eq.~\eqref{eq:Hessian_EMA}. Here we use \nameo(prefix) to fine-tune Roberta-large on SNLI dataset($K=512$). More explored experiments can be found in Appendix \ref{app:smooth_scale}. }
% \label{fig:smooth_scale}
% %\vskip -0.15in
% \end{figure}

\subsection{Hyperparameter Analysis}
\label{Hyperparameter Analysis}

\vspace{-7pt}
\begin{figure}[h]
    %\centering
    \hspace{0.1in}
    \begin{minipage}[t]{0.46\textwidth}
      \centering
      \includegraphics[width=1\linewidth]{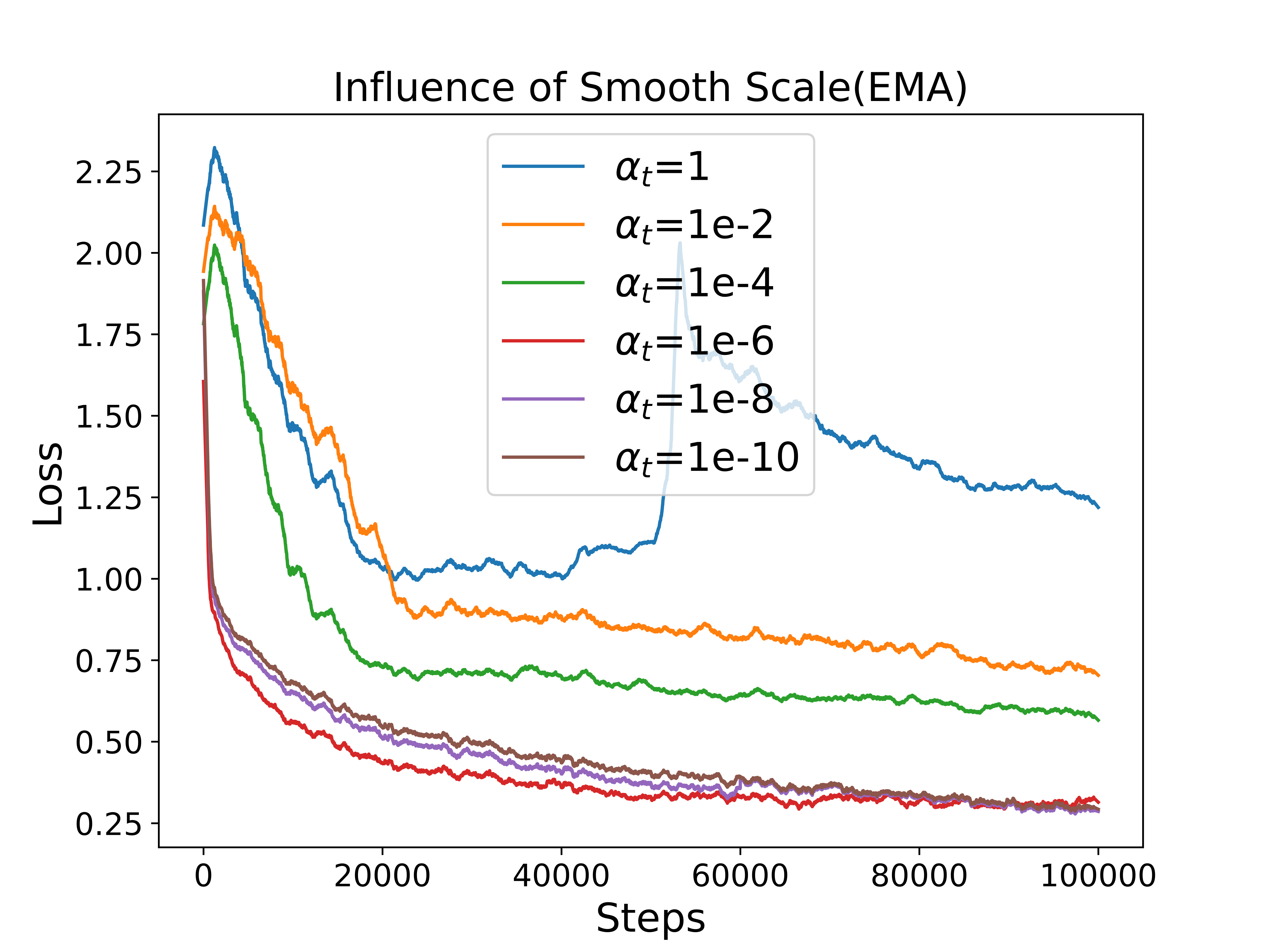}
      \vspace{-16pt}
      \caption{Influence of EMA $\alpha_t$ for hessian in Eq.~\eqref{eq:Hessian_EMA}. We use \nameo(prefix) to fine-tune Roberta-large on SNLI. More results can be found in Appendix \ref{app:smooth_scale}.
      }
      \vspace{-3pt}
      \label{fig:smooth_scale}
    \end{minipage}
    %\hfill
    \hspace{0.1in}
    \begin{minipage}[t]{0.46\textwidth}
      \centering
      \includegraphics[width=0.9\linewidth]{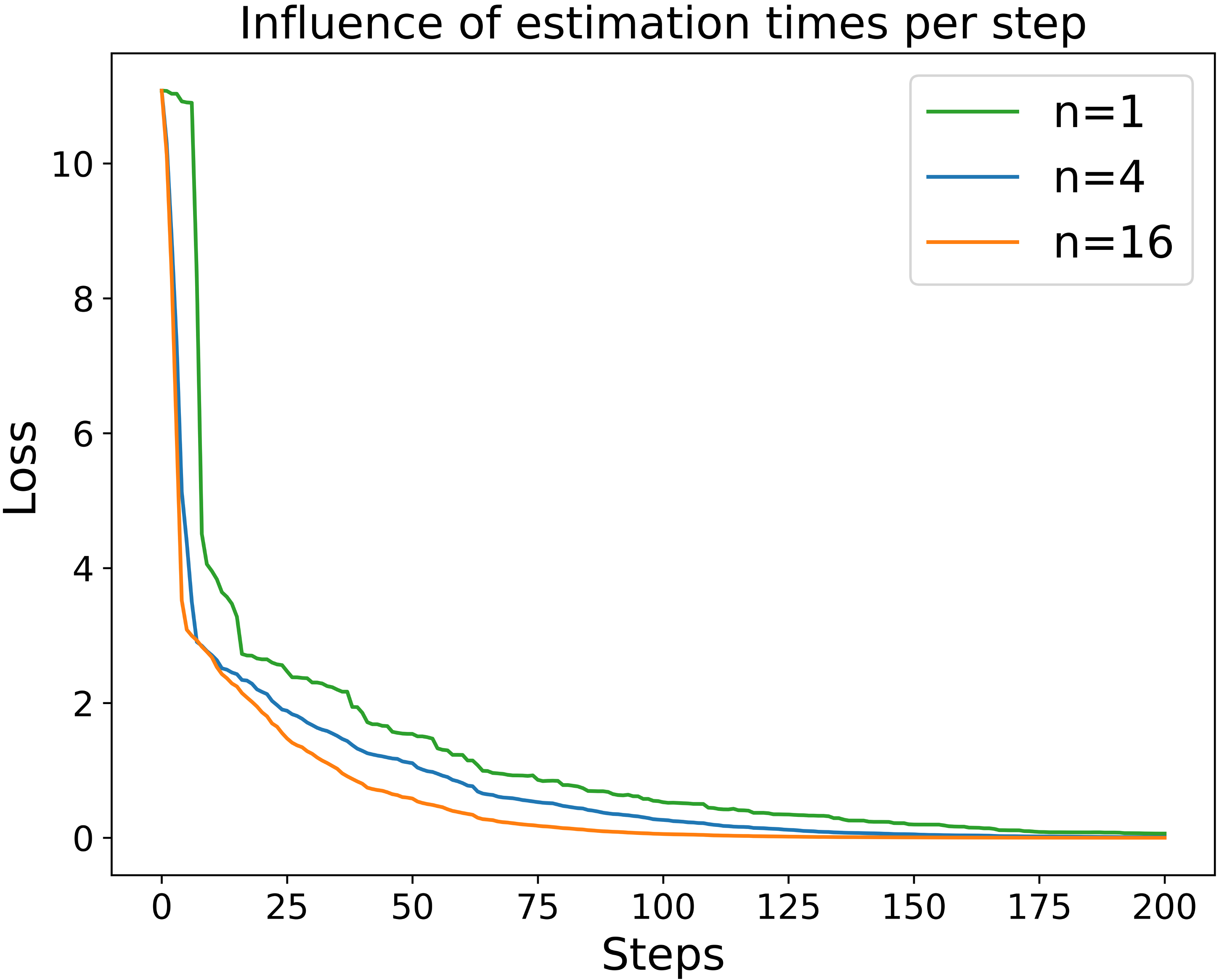}
      \vspace{-5pt}
      \caption{Loss curves on Function (a) using the variant HiZOO-multi with different estimation times $n$ per step. Trajectory visualization can be found in Appendix~\ref{app:smooth_scale}.
      }
      \vspace{-4pt}
      \label{fig:estimation_times}
    \end{minipage}
    
\end{figure}
\vspace{-2pt}

\textbf{Influence of Smooth Scale $\alpha_t$ in EMA}~~~~To assess the robustness of the optimizer, a grid search is conducted to evaluate the sensitivity of the hyper-parameter $\alpha_t$ on RoBERTa-large (350M). Figure~\ref{fig:smooth_scale} illustrates that % the smoothing scale $\alpha_t$ is instrumental in controlling the convergence rate during training. 
as $\alpha_t$ is incrementally increased from zero, the training loss decreases faster. However, too large $\alpha_t$ values may impede convergence or even cause training to fail due to gradient explosion.% The experimental results indicate that $\alpha_t = 1e-6$ yields better performance in most cases.

\textbf{Influence of Estimation Times $n$ Per Step}~~~~We also propose a variant of HiZOO in Appendix~\ref{HiZOO_multi}: HiZOO-multi, which has $n\textgreater1$ per step. As shown in Figure~\ref{fig:estimation_times}, different $n$ maybe doesn't affect the final accuracy. However, the larger $n$ will estimate the diagonal Hessian more accurate per step and accelerate model convergence, reducing the overall training steps. But it will also increase the computation per step. Balancing these factors is crucial for efficient training.

\section{Conclusion}
\label{sec_conslusion}

In this work, we introduce \name, which is the first ZOO that incorporates diagonal Hessian for fine-tuning LLMs. By introducing one more forward pass, HiZOO can handle heterogeneous curvatures across different parameter dimensions. We provide theoretical analysis and visualize the optimization trajectories to explore how it works. Further experiments show that \nameo converges in much fewer steps than MeZO and achieves better performance across various LLMs. We also explore a memory efficient implementation (HiZOO-L) to reduce the Hessian consumption. %In the future, we and there are also many other methods worth trying, such as quantization~\citep{dettmers2023qlora}, FlashAttention~\citep{dao2022flashattention}. %As a limitation, HiZOO 

%\section*{Impact Statements}
%This paper presents work whose goal is to advance the field of Machine Learning. There are many potential societal consequences of our work, none which we feel must be specifically highlighted here.

%\newpage
%\usepackage[style=authoryear]{biblatex}
%\bibliographystyle{abbrvnat}
%\bibliographystyle{plain}
\bibliographystyle{iclr2025_conference}
\bibliography{sections/7_paper_ref}

\newpage
\appendix
\section{Related Works}
\label{app_related_works}

\subsection{First-order Optimizer Used in LLMs}

Optimization methods have consistently been a popular research domain, encompassing techniques such as Gradient Descent (GD), Momentum, Adagrad \citep{Adagrad}, ADADELTA~\citep{zeiler2012adadelta}, and Newton's method, which have been instrumental in advancing fields like computer vision. However, the emergence of large-scale models, characterized by their massive parameter counts and intricate architectures, has challenged the efficacy of conventional optimization methods for training tasks. Amidst this landscape, Adam \citep{Adam} has emerged as the preferred choice for its ability to rapidly converge, making it particularly suitable for the training and fine-tuning large models. Then AdamW \citep{adamw} was proposed to add a weight decay coefficient to alleviate over-fitting. %This variant has been broadly applied across a spectrum of pre-trained language models. 
Notwithstanding these advancements, a limitation persists with these optimizers: they have an implicit batch size ceiling. Exceeding this threshold can provoke extreme gradient updates, thus impeding the convergence rate of the models. This bottleneck is particularly problematic in the context of large-model training, which typically necessitates substantial batch sizes. To circumvent this constraint, LAMB \citep{LAMB} was devised to apply principled layer-wise adaptation strategy to accelerate the training of large models employing large batches.

\subsection{Hessian Based First-Order Optimizer}

Compared with first-order optimizers, second-order optimizer considers second-order information  in the process of gradient calculation. As a result, it has more abundant information to guide gradient descent and is considered to be more promising. Previous studies utilized curvature information to pre-condition the gradient~\citep{second-order-curvature-information1, second-order-curvature-information2, second-order-curvature-information3}. Subsequently, Magoulas et al.~\citep{diagonal_Hessian1} applied diagonal Hessian as the pre-conditioner, which greatly promotes the landing of second-order optimizer in the field of deep learning. Martens \citep{conjugate_gradient} approximated the Hessian with conjugate gradient. Schaul et al.~\citep{Schaul_tuneLR_usingDiagonalHessian} utilized diagonal Hessian to automatically adjust the learning rate of SGD during training. Another work~\citep{Gaussian-Newton-approximation-Hessian} extended natural gradient descent to incorporate second order information alongside the manifold information and used a truncated Newton approach for inverting the metric matrix instead of using a diagonal approximation of it. EVA ~\citep{nn_approximate_Hessian4} proposed to use the Kronecker factorization of two small vectors to approximated the Hessian, which significantly reduces memory consumption. AdaHessian~\citep{ADAHESSIAN} incorporates an approximate Hessian diagonal, with spatial averaging and momentum to precondition the gradient vector.

Although great progress has been made in the research of second-order optimizer, it has not been widely used because of the extra computation and memory cost when gradient updating, and this situation is extremely serious in the training of large language models. Based on the above dilemma, recent works~\citep{anil2021scalable, George18} proposed to offload Hessian computation to CPUs and utilized ResNets and very large batch size to approximate the Fisher information matrix. Sophia~\citep{Liu2023sophia:} was the first to apply second-order optimizer and achieve a speed-up on large language models in total compute successfully.

\subsection{Zeroth-Order Optimizer}

Zeroth-order optimization, is also known as derivative-free or black-box optimization. There have been many one-point gradient estimators in past works~\citep{FairScale2021, Stochastic_Approximation, Stochastic_Approximation2, ZO_GradientEstimator_SPSA, ZO1, Zo2, ZO3, wang2020zeroth}. However, cursory experiments with one such promising estimator \citep{feedback_black_box} reveal that SPSA outperforms other methods.

In previous works, it appears in a wide range of applications where either the objective function is implicit or its gradient is impossible or too expensive to compute. For example, methods~\citep{distributed_ZO1, distributed_ZO2} consider derivative-free distributed algorithms for non-convex multi-agent optimization. ZO-BCD\citep{black_box_adversial_ZO2}, ZOO\citep{black_box_adversial_ZO1}, ZO-signSGD \citep{black_box_adversial_ZO3} and ZO-HessAware \citep{black_box_adversial_ZO4} utilize zeroth-order stochastic optimization to generate black-box adversarial example in deep learning.

Beyond that, MeZO \citep{mezo} firstly adapted the classical ZO-SGD method to fine-tune LLMs, while achieving comparable performance with extremely great memory reduction and GPU-hour reduction. Recently there have been many subsequent excellent works~\citep{jiang2023zoadamu, zhao2024helene, sparseMeZO, MeZOsparsity, tang2024effectivelyzo, chen2024zo_lowrank, wangadvancementzo, chen2025towardszo, tan2025harmonyzo, chen2025memoryzo, sun2025tezo}. All of these optimizers provide researchers with a new and promising technique for fine-tuning large models.

%Zeroth-order optimization Many classical lower bounds have been derived for ZO-SGD in the strongly convex and convex settings [47, 3, 79, 32, 85, 69] as well as non-convex [101]. These bounds generally depended on the number of parameters d. More recently, [100, 6, 15] showed that if the gradient has low-dimensional structure, then the query complexity scales linearly with the intrinsic dimension and logarithmically with the number of parameters, though the estimation has at least Ω(sd log d) memory cost. Additional tricks such as sampling schedules [11] and other variance reduction methods [48, 62] can be added to ZO-SGD. ZO has inspired distributed methods [93, 43] and black-box adversarial example generation [14, 63, 17, 64] in deep learning. Ye et al. [108], Balasubramanian and Ghadimi [7] estimate the Hessian to perform ZO optimization along important directions. There are also ZO methods that optimize without estimating the gradient [38, 68, 44].

\section{Detailed Convergence Analysis}
\label{app_convergence_proof}

Firstly, our convergence analysis requires the following assumptions:
\begin{assumption}\label{ass:L}
    The objective function $\mL(\theta)$ is $L$-smooth, which means that for any $\theta_1, \theta_2 \in \RR^d$, it holds that:
    \begin{equation}
        \mL(\theta_2) \leq \mL(\theta_1) + \langle \nabla \mL(\theta_1), \theta_2 - \theta_1 \rangle + \frac{L}{2}\|\theta_2 - \theta_1\|^2.
    \end{equation}
\end{assumption}
\begin{assumption}\label{ass:sigma}
    The stochastic gradient $\nabla \mL(\theta)$ has $\sigma^2$ variance, which means: 
    \begin{equation}
        \EE\left[ \|\nabla \mL(\theta) - \nabla \mL(\theta)\|^2 \right] \leq \sigma^2.
    \end{equation}
\end{assumption}
\begin{assumption}\label{ass:beta}
    Each entry of $\Sigma_t$ lies in the range $[\beta_\ell, \beta_u]$ with $0<\beta_\ell \leq \beta_u$.
\end{assumption}

Then we will give the detailed proof for convergence.

\begin{proof}
By the update rule of $\theta_t$ and Assumption~\ref{ass:L}, we have
\begin{align*}
&\EE\left[\mL(\theta_{t+1}) \mid \theta_t\right] \\
\leq&
\mL(\theta_t) - \eta_t \EE\left[\langle \nabla \mL(\theta_t), g_\mu(\theta_t) \rangle\right] + \frac{L\eta_t^2}{2}\EE\left[\|g_\mu(\theta_t)\|^2\right]\\
\leq&
\mL(\theta_t) - \eta_t \|\nabla \mL(\theta_t)\|_{\Sigma_t}^2 +  \eta_t \cO \left(\mu \|\nabla \mL(\theta_t) \|\right)\\
&+
2\eta_t^2L\left( \mathrm{tr}(\Sigma_t) +\beta_u \right) \|\nabla \mL(\theta_t)\|_{\Sigma_t}^2 \\
&+ 2\eta_t^2L\left( \mathrm{tr}(\Sigma_t) +\beta_u \right)\sigma^2 + \cO(\mu^2)\\
\leq&
\mL(\theta_t) - \frac{\eta_t}{2} \|\nabla \mL(\theta_t)\|_{\Sigma_t}^2 
+
2\eta_t^2L\left( \mathrm{tr}(\Sigma_t) +\beta_u \right) \|\nabla \mL(\theta_t)\|_{\Sigma_t}^2 \\
&+ 2\eta_t^2L\left( \mathrm{tr}(\Sigma_t) +\beta_u \right)\sigma^2 + \cO(\mu^2)\\
=&
\mL(\theta_t) - \frac{\eta_t}{2}\left(1 - 4\eta_t L( \mathrm{tr}(\Sigma_t) +\beta_u )\right)\|\nabla \mL(\theta_t)\|_{\Sigma_t}^2\\ 
&+ 2\eta_t^2L\left( \mathrm{tr}(\Sigma_t) +\beta_u \right)\sigma^2 + \cO(\mu^2)\\
\leq&
\mL(\theta_t) - \frac{\eta_t}{4} \|\nabla \mL(\theta_t)\|_{\Sigma_t}^2  + 2\eta_t^2L\left( \mathrm{tr}(\Sigma_t) +\beta_u \right)\sigma^2 + \cO(\mu^2),
\end{align*}
where the second inequality is because of Lemma~\ref{lem:gmu} and the last inequality is because of the value of $\eta_t$.

Rearrange above equation and summing up it, we can obtain that 
\begin{align*}
   &\EE\left[ \sum_{t=1}^T\frac{\eta_t}{4} \|\nabla \mL(\theta_t)\|_{\Sigma_t}^2\right] 
   \leq \sum_{t=1}^T \left( \mL(\theta_t) - \mL(\theta_{t+1}) \right)  \\
   &+  2\eta_t^2L\left( \mathrm{tr}(\Sigma_t) +\beta_u \right)\sigma^2 + \cO(T\mu^2)\\
   =&
   \mL(\theta_1) - \mL(\theta_{T+1}) +  2\eta_t^2L\left( \mathrm{tr}(\Sigma_t) +\beta_u \right)\sigma^2 + \cO(T\mu^2)\\
   \leq&
   \mL(\theta_1) - \mL(\theta_*) +  2\eta_t^2L\left( \mathrm{tr}(\Sigma_t) +\beta_u \right)\sigma^2 + \cO(T\mu^2).
\end{align*}

By taking $\theta_{\mbox{out}} = \theta_j$ with $j$ uniformly sampled from $\{1, \dots, T\}$ and taking expectation, we can obtain that
\begin{align*}
    &\EE\left[ \|\nabla \mL(\theta_{\mbox{out}})\|^2\right] 
    =
    \frac{1}{T} \sum_{t=1}^T \|\nabla \mL(\theta_t)\|^2 \leq \frac{1}{T \beta_\ell} \sum_{t=1}^T \|\nabla \mL(\theta_t)\|_{\Sigma_t}^2\\
    \leq&
    \frac{4 (\mL(\theta_1) - \mL(\theta_*))}{T\beta_\ell \eta} + \frac{8\eta L\left( \mathrm{tr}(\Sigma_t) +\beta_u \right)}{T\beta_\ell} \sigma^2 + \cO(\mu^2)\\
    =&
    \frac{32L\left( \mathrm{tr}(\Sigma_t) +\beta_u \right) (\mL(\theta_1) - \mL(\theta_*))}{\sqrt{T}\beta_\ell } + \frac{\sigma^2}{T^{3/2} \beta_\ell} + \cO\left(\mu^2\right),
\end{align*}
where the first inequality is because of the assumption that the diagonal entries of $\Sigma_t$ is no less than $\beta_\ell$,
\end{proof}

Eq.~\eqref{eq:theta} shows that once we choose the step size $\eta$ properly, $\mL(\theta_{t+1})$ will be less than $\mL(\theta_t)$ in expectation up to some noises of order $\mu^2$. 
Specifically, if set $\eta = \frac{1}{8\sqrt{T}L(\max_t\mathrm{tr}(\Sigma_t) +\beta_u)}$, Eq.~\eqref{eq:Lout} implies that we can find an solution $\theta_{\mbox{out}}$ such that $\EE\left[\|\nabla \mL(\theta_{\mbox{out}})\|^2\right] \leq \epsilon^2$ in $\cO(\epsilon^{-4})$ iterations.
This rate matches the one of \citep{ghadimi2013stochastic}.

\begin{lemma}\label{lem:gmu}
We assume that Assumption~\ref{ass:sigma} and Assumption~\ref{ass:beta} hold. 
Then, $g_\mu(\theta_t)$ defined in Eq.~\eqref{eq:gmu} has the following properties:
\begin{align*}
     &\EE\left[ g_\mu(\theta_t) \right] 
        = \Sigma_t \nabla \mL(\theta_t) + \cO(\mu)\\
    &\EE\left[ \|g_\mu(\theta_t)\|^2\right]
        \leq
        4\left( \mathrm{tr}(\Sigma_t) + \beta_u  \right) \|\nabla \mL(\theta_t)\|_{\Sigma_t}^2 \\
       & + 4\beta_u\left( \mathrm{tr}(\Sigma_t) + \beta_u  \right)\sigma^2 + \cO(\mu^2).
\end{align*}
\end{lemma}
\begin{proof}
By the definition of $g_\mu(\theta_t)$, we have
    \begin{align*}
        &g_\mu(\theta_t) \\
        = &
        \sum_{i=1}^b \frac{\mL(\theta_t + \mu \Sigma_t^{1/2} u_i) - \mL(\theta_t - \mu \Sigma_t^{1/2} u_i)}{2b\mu} \Sigma_t^{1/2}u_i\\
        =&
        \sum_{i=1}^b \frac{2\mu \nabla^\top \mL(\theta_t) \Sigma_t^{1/2} u_i + \cO(\mu^2)}{2b\mu} \Sigma_t^{1/2}u_i \\
        =&
        \frac{1}{b}\sum_{i=1}^b \Sigma_t^{1/2}u_i u_i^\top \Sigma_t^{1/2} \nabla \mL(\theta_t) + \cO(\mu).
    \end{align*}
    Thus, we can obtain that
    \begin{align}
        \EE\left[ g_\mu(\theta_t) \right] 
        = \Sigma_t \nabla \mL(\theta_t) + \cO(\mu).
    \end{align}
    Moreover,
    \begin{align*}
        &\EE\left[ \|g_\mu(\theta_t)\|^2\right]\\
        =&
        \EE\bigg[ \| \frac{1}{b}\sum_{i=1}^b \Sigma_t^{1/2}u_i u_i^\top \Sigma_t^{1/2} \nabla \mL(\theta_t) + \cO(\mu)  \|^2 \bigg]\\
        \leq&
        2\EE\left[\| \frac{1}{b}\sum_{i=1}^b \Sigma_t^{1/2}u_i u_i^\top \Sigma_t^{1/2} \nabla \mL(\theta_t)
         \|^2\right] + \cO(\mu^2)\\
        \leq&
        \frac{2}{b}\sum_{i=1}^b\EE\left[ \| \Sigma_t^{1/2}u_i u_i^\top \Sigma_t^{1/2} \nabla \mL(\theta_t) \|^2 \right]+ \cO(\mu^2)\\
        =&
        2 \mathrm{tr}(\Sigma_t) \cdot \nabla^\top \mL(\theta_t) \Sigma_t \nabla \mL(\theta_t) \\
        &+ 2\nabla^\top \mL(\theta_t) \Sigma_t^2 \nabla \mL(\theta_t) + \cO(\mu^2)\\
        \leq&
        2\left( \mathrm{tr}(\Sigma_t) + \beta_u  \right) \nabla^\top \mL(\theta_t) \Sigma_t \nabla \mL(\theta_t) + \cO(\mu^2),
    \end{align*}
    where  the last equality is because of Lemma~\ref{lem:1}.

    Finally, we have
    \begin{align*}
        &\EE\left[\nabla^\top \mL(\theta_t) \Sigma_t \nabla \mL(\theta_t) \right]
        = \EE\left[\|\nabla \mL(\theta_t)\|_{\Sigma_t}^2 \right]\\
        \leq&
       2\EE\left[\|\nabla \mL(\theta_t)\|_{\Sigma_t}^2\right] +  2\EE\left[\| \nabla \mL(\theta_t) - \nabla \mL(\theta_t) \|_{\Sigma_t}^2\right]\\
       \leq&
       2\|\nabla \mL(\theta_t)\|_{\Sigma_t}^2 + 2\beta_u \EE\left[ \| \nabla \mL(\theta_t) - \nabla \mL(\theta_t) \|^2 \right]\\
       \leq&
       2\|\nabla \mL(\theta_t)\|_{\Sigma_t}^2 + 2\beta_u \sigma^2,
    \end{align*}
    where the second inequality is because of Assumption~\ref{ass:beta} and the last inequality is because of Assumption~\ref{ass:sigma}.

    Therefore,
    \begin{align*}
        &\EE\left[ \|g_\mu(\theta_t)\|^2\right]
        \leq
        4\left( \mathrm{tr}(\Sigma_t) + \beta_u  \right) \|\nabla \mL(\theta_t)\|_{\Sigma_t}^2 
        + 4\beta_u\left( \mathrm{tr}(\Sigma_t) + \beta_u  \right)\sigma^2.
    \end{align*}
\end{proof}

\begin{lemma}\citep{magnus1978moments}\label{lem:1}
	Let $A$ and $B$ be two symmetric matrices, and $u$ obeys the Gaussian distribution, that is, $u \sim N(0,I_d)$. Define $z = u^\top A u\cdot u^\top B u$. The expectation of $z$ is:
	{\begin{equation}\label{eq4}
			\small
			\begin{aligned}
				&\mathbb{E}_u[z] =  (\mathrm{tr}A)(\mathrm{tr}B) + 2\mathrm{tr} (AB).
			\end{aligned}
		\end{equation}
	}
\end{lemma}

\section{Test Functions of the optimization trajectories}
\label{app_test_functions}

\begin{figure}[h]
\centering
\scalebox{1.1}{
\includegraphics[width=0.9\linewidth]{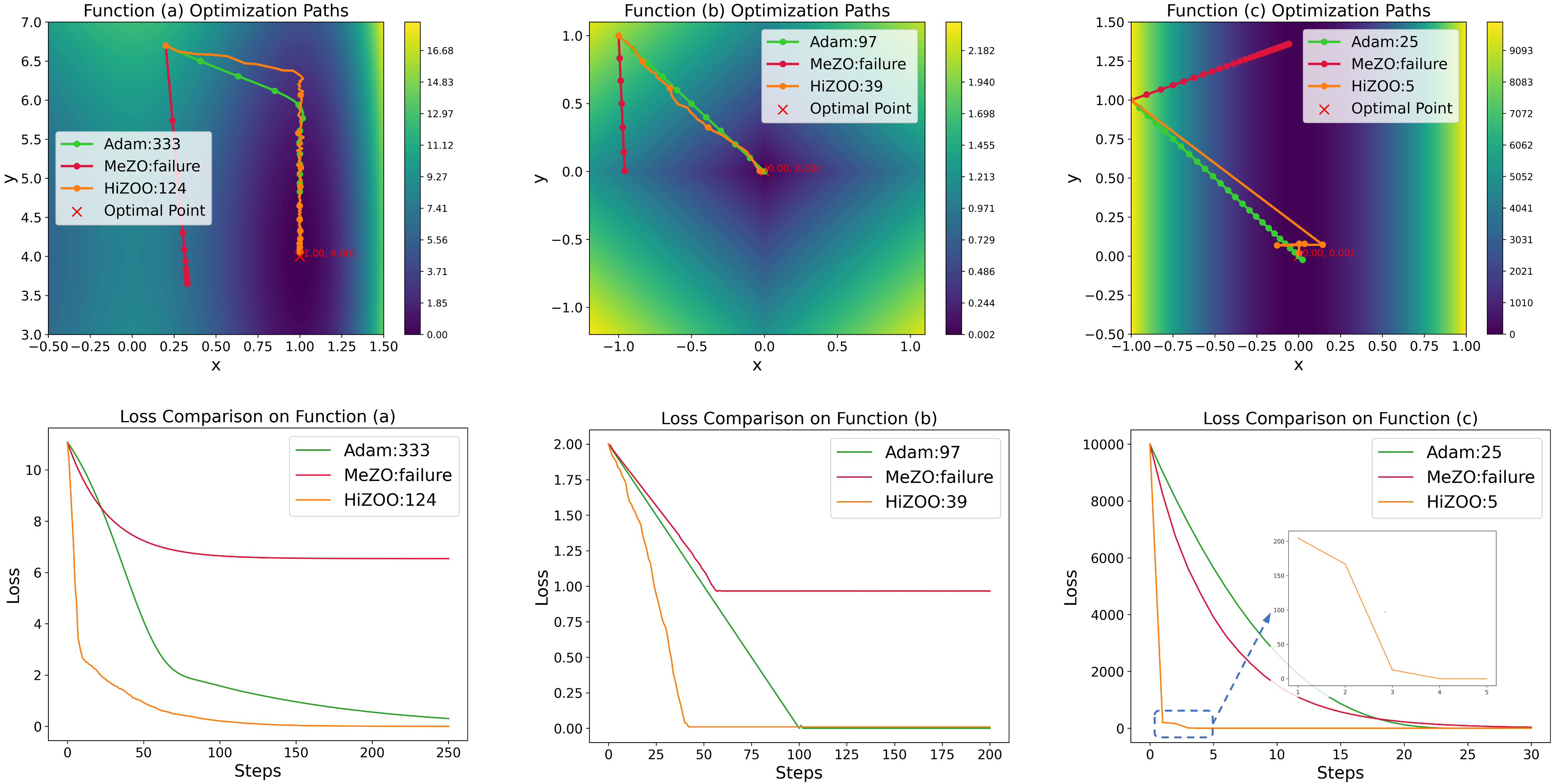}}
\vspace{-3mm}
\caption{2D trajectories of Adam, MeZO and HiZOO on 3 test functions. The upper figures are the 2D trajectories of gradient descent, and the bottom parts are the corresponding loss curves.}
\label{fig:fun_loss}
%\vskip -0.2in
\end{figure}

For better illustrating how HiZOO utilizes hessian to improve the convergence process, we choose below three test functions with heterogeneous curvatures across different parameters. In Figure~\ref{fig:fun_loss}, we provide the 2D convergence paths of three functions and the variation of their losses with respect to steps.

\begin{itemize}
\setlength{\itemsep}{1pt}
\vspace{-3mm}
\item Function (a)\footnote{Function (a) is from \citep{Liu2023sophia:}.}: $f(x,y) = 8(x-1)^2(1.3x^2+2x+1)+0.5(y-4)^2$
\item Function (b): $f(x,y) = \lvert x \rvert + \lvert y \rvert$
\item Function (c): $f(x,y) = 10000 x^2 + y^2$
\vspace{-3mm}
\end{itemize}

\section{HiZOO Variants}
\label{app_MeZO_Hessian_variants}

\subsection{HiZOO-L}
\label{HiZOO_L}

\begin{figure}[h]
% \vskip -0.1in
\centering
\scalebox{1.0}{
\includegraphics[width=1.0\linewidth]{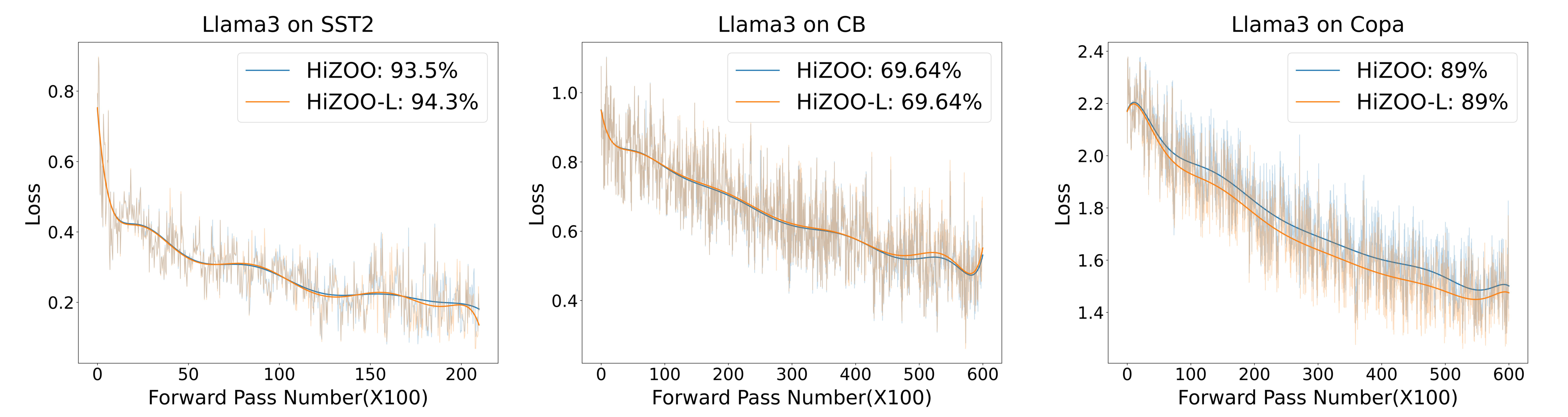}}
\vspace{-3mm}
\caption{Loss curves on Llama3 between HiZOO and HiZOO-L.}
\label{fig:ada_llama_loss}
%\vskip -0.2in
\end{figure}

Due to the storage of Hessian, HiZOO introduces extra memory cost, which is equal to the size of the model parameters. To address this limitations, we propose HiZOO-L, the low-rank implementation for the storage of Hessian, motivated by Adafactor~\citep{adafactor}. Details can be see in Algorithm~\ref{alg:HiZOO_adafactor}. %We conduct lots of experiments to show the performance of HiZOO-L, as results shown in Table~\ref{tab:adafactor} and Figure~\ref{fig:ada_opt_llama}. 
We also visualize the loss curves of HiZOO and HiZOO-L in Figure~\ref{fig:ada_llama_loss} and find that on most datasets two algorithms perform closely. This also indicates that the estimation of Hessian in HiZOO may be sparse, so we encourage researchers to try other memory efficient algorithms to compress the Hessian.

\begin{algorithm}[h]
\begin{algorithmic}[1]
%\setstretch{1.05}

  %\KwIn{Initial parameter $\theta$, perturbation size $\epsilon$, and number of iterations $L$}
  %\KwOut{Optimal parameter $\theta^*$}
  \Require parameters $\theta \in \mathbb{R}^d$, loss $L : \mathbb{R}^d \rightarrow \mathbb{R}$, step budget $T$, perturbation scale $\mu$, learning rate schedule ${\eta_t}$, smooth scale $\alpha_t$, diagonal Hessian $R_0$, $C_0$

  %\vspace{0.1cm}
  \For{$t=1,...,T$} 
    \State Sample batch $ \mathcal{B} \subset \mathcal{D}$ and random seed $s$
    \State $\ell \leftarrow \mathcal{L}(\theta; \mathcal{B})$ 
    \State $\theta$ $\leftarrow$ PerturbParameters($\theta$, $\mu$,  $R_{t-1}$, $C_{t-1}$, $s$)
    \State $\ell_{+} \leftarrow \mathcal{L}(\theta; \mathcal{B})$
    \State $\theta$ $\leftarrow$ PerturbParameters($\theta$, $-2 \mu$, $R_{t-1}$, $C_{t-1}$, $s$)
    \State $\ell_{-} \leftarrow \mathcal{L}(\theta; \mathcal{B})$
    %\vspace{1mm}\State \CommentSty{//Reset parameters before descent}
    \State $\theta$ $\leftarrow$ PerturbParameters($\theta$, $\mu$, $R_{t-1}$, $C_{t-1}$, $s$)
    \Comment{Reset parameters before descent}
    
    \State $\hat{\Sigma}^{-1}_{t-1} = (R_{t-1} * C_{t-1})/(1_{n}^{\top}*R_{t-1})$
    \State $\hat{\Sigma}^{\prime}_{t} = \frac{1}{2 \mu ^2 } (\ell_{+} + \ell_{-} - 2 \ell )(\hat{\Sigma}^{-1/2}_{t-1} u_i u_i ^{\top} \hat{\Sigma}^{-1/2}_{t-1})$

    \State $R_t^{-1} = (1-\alpha_t)R_{t-1}^{-1} + \alpha_t \left| diag(\hat{\Sigma}^{\prime}_{t})\right| * 1_m$
    \State $C_t^{-1} = (1-\alpha_t)C_{t-1}^{-1} + \alpha_t 1^{\top}_n * \left|   diag(\hat{\Sigma}^{\prime}_{t})\right|$
    \State projected\_grad $\leftarrow (\ell_{+} - \ell_{-}) \ast \hat{\Sigma}^{1/2}_t /2\mu$
    
    %\State \CommentSty{//For sampling $u_i$}
    \State Reset random number generator with seed $s$ 
    \Comment{For sampling $u_i$}
    %\State  \CommentSty{//Update parameters in place}
    \For{$\theta_i \in \theta$}
        \State Sample $u_i \sim \mathcal{N}(0,I_d)$
        \State $\theta_i \leftarrow \theta_i - \eta_t \ast$  projected\_grad $\ast u_i$
    \EndFor
  \EndFor
  \Statex
  \Function {PerturbParameter}{$\theta$, $\mu$, $R_{t}$, $C_{t}$, $s$}
    %\State \CommentSty{//For sampling $u_i$}
    \State Reset random number generator with seed $s$  
    \Comment{For sampling $u_i$} 
    %\State  \CommentSty{//Modify parameters in place}
    \For{$\theta_i \in \theta$}
        \vspace{1mm}\State Sample $u_i \sim \mathcal{N}(0,I_d)$
        \State $\hat{\Sigma}^{-1}_t = (R_t * C_t)/(1_{n}^{\top}*R_t)$
        \State $\theta_i \leftarrow \theta_i + \mu \hat{\Sigma}^{1/2}_t u_i$
        \Comment{Modify parameters in place}
    \EndFor
    \State \Return $\theta$

  \EndFunction
  \caption{\name-L}

  \label{alg:HiZOO_adafactor}
\end{algorithmic}
\end{algorithm}

\subsection{HiZOO-multi}
\label{HiZOO_multi}

There is a rich history of transferring ideas from first order optimization to enhance ZO algorithms. Below, we highlight the variant of \name: \name-multi which can perform $n$ estimation times per step  efficiently as shown in Algorithm \ref{alg:MeZO_Hessian_multi}. We conducted experiments to explore the influence of estimation times $n$ per step as shown in Figure~\ref{fig:app_estimation_times}. We can conclude that when $n$ is larger, the estimation of diagonal Hessian is more accurate. It can decrease the variance of the estimated diagonal Hessian matrix during each step and thus reduce the overall training steps, but will cause much more computation per step meanwhile. So choosing an appropriate value of $n$ is very important during the training.

%\input{appendix/app_adafactor}
% \begin{figure}[h]
% % \vskip -0.1in
% \centering
% \scalebox{1.0}{
% \includegraphics[width=1.0\linewidth]{appendix/adafactor_opt_llama3.png}}
% % \vspace{-3mm}
% \caption{Accuracy comparison on OPT-13B (left) and Llama3 (right) between HiZOO and HiZOO-L.}
% \label{fig:ada_opt_llama}
% %\vskip -0.2in
% \end{figure}

\begin{algorithm}[h] % {1.0\textwidth}
%\centering
\begin{algorithmic}[1]
\Require parameters $\theta \in \mathbb{R}^d$, loss $L : \mathbb{R}^d \rightarrow \mathbb{R}$, step budget $T$, perturbation scale $\mu$, batch size $B$, learning rate schedule ${\eta_t}$, smooth scale $\alpha_t$, estimate times $n$, Hessian matrix $\Sigma_{0}$
  
%\vspace{0.2cm}
\For{$t=1,...,T$} 
\State seeds, projected\_grads $\leftarrow$ [] 
\For{$j=1,...,n$} 
    \State Sample batch $\mathcal{B} \subset \mathcal{D}$ and random seed $s$
    \State $\ell \leftarrow \mathcal{L}(\theta; \mathcal{B})$ 
    \State $\theta$ $\leftarrow$ PerturbParameters($\theta$, $\mu$, $\Sigma^{1/2}_{t-1}$, $s$)
    \State $\ell_{+} \leftarrow \mathcal{L}(\theta; \mathcal{B})$
    \State $\theta$ $\leftarrow$ PerturbParameters($\theta$, $-2 \mu$, $\Sigma^{1/2}_{t-1}$, $s$)
    \State $\ell_{-} \leftarrow \mathcal{L}(\theta; \mathcal{B})$ 
    \State $\theta$ $\leftarrow$ PerturbParameters($\theta$, $\mu$, $\Sigma^{1/2}_{t-1}$, $s$) \Comment{Reset parameters before descent}
    \State $\Sigma^{\prime}_{t} = \frac{1}{2 \mu ^2 } (\ell_{+} + \ell_{+} - 2 \ell )(\Sigma^{-1/2}_{t-1} u_i u_i ^{\top} \Sigma^{-1/2}_{t-1})$
        %\State \CommentSty{//Update Hessian matrix}\\
    \State $\Sigma_{t}^{-1} = (1-\alpha_t) \Sigma_{t-1}^{-1} + \alpha_t \left| diag(\Sigma^{\prime}_{t}) \right|$
    \Comment{Update Hessian matrix}
    \State projected\_grad $\leftarrow$ $(\ell_+ - \ell_-) * \Sigma^{1/2}_{t}/2 \mu $ 
\State projected\_grads[j] $\leftarrow$ projected\_grad 
\State seeds[j] $\leftarrow s$
\EndFor
\EndFor

\For{$j=1,...,n$} 
\State Reset random number generator with seeds[j] 
\For{$\theta_i \in \theta$}
\State $u_i \sim \mathcal{N}(0,I_d)$ 
\State $\theta_i \leftarrow \theta_i - (\eta_t/n)*$projected\_grads[j] $* u_i$ 
\Comment{Avg grad for $u_1,...,u_n$ }
\EndFor
\EndFor

\Statex
\Function{PerturbParameter} {$\theta$, $\mu$, $\Sigma^{1/2}_t$,$s$}
\State Reset random number generator with seed $s$
\For{$\theta_i\in\theta$ }  
    \State $u_i \sim \mathcal{N}(0,I_d)$
    \State $\theta_i \leftarrow \theta_i + \mu \Sigma^{1/2}_t u_i$ \Comment{Modify parameters in place}
\EndFor
\State \Return $\theta$

\EndFunction

\caption{\name-multi}
\label{alg:MeZO_Hessian_multi}
\end{algorithmic}
\end{algorithm}

\section{Experiments on LLMs}

\subsection{Detailed Experiments on RoBERTa-large}
\label{appendix_robert}

%\subsubsection{Experiment Hyperparameters}
%\label{app:roberta_hyperpara}
We use the hyperparameters in Table \ref{tab:roberta_hyper} for \nameo experiments on RoBERTa-large. Regarding learning rate scheduling and early stopping, we use constant learning rate for all \nameo experiments. %For RoBERTa-large experiments, we evaluate the model on validation sets every 1/10 of total training steps and save the best validation checkpoint.

\begin{table*}[h]
\centering
\caption{The hyperparameter grids used for RoBERTa-large experiments. \nameo uses a constant learning rate schedule. All \nameo experiments use $100$K steps.}
\small
% \resizebox{0.8\columnwidth}{!}{
\begin{tabular}{lrc}
\toprule
Experiment & Hyperparameters & Values \\
\midrule
\name & Batch size & $64$ \\
& Learning rate & $\{1\mathrm{e}{-7}, 1\mathrm{e}{-6}, 1\mathrm{e}{-5} \}$ \\
& $\mu$ & $1\mathrm{e}{-3}$ \\
& Weight Decay & $0$ \\
\midrule
\name (prefix) & Batch size & $64$ \\
& Learning rate & $\{1\mathrm{e}{-2}, 5\mathrm{e}{-3}, 1\mathrm{e}{-3} \}$ \\
& $\mu$ & $1\mathrm{e}{-1}$ \\
& Weight Decay & $0$ \\
& \# prefix tokens &$5$\\
\midrule
\name (LoRA) & Batch size & $64$ \\
& Learning rate & $\{1\mathrm{e}{-5}, 5\mathrm{e}{-5}, 1\mathrm{e}{-4} \}$ \\
& $\mu$ & $1\mathrm{e}{-3}$ \\
& Weight Decay & $0.1$ \\
& $(r, \alpha)$ & $(8, 16)$\\
%\midrule
\bottomrule
\end{tabular}
% }

\label{tab:roberta_hyper}
\end{table*}
\begin{table*}[htbp]
\vspace{-8pt}
  \centering
  \caption{Experiments on RoBERTa-large (350M parameters, k=512). For MeZO we report the results we reproduced.}% that include zero-shot learning, linear probing (LP), FT with Adam, MeZO and HiZOO, and PEFT (LoRA, prefix ) with Adam, MeZO and HiZOOo respectively. All reported numbers are averaged accuracy (standard deviation) over 5 runs.}
    \vspace{2pt}
    \scalebox{0.8}{
    \begin{tabular}{lccccccc}
    \toprule
    Task Type & \multicolumn{1}{c}{\textbf{SST-2}} & \multicolumn{1}{c}{\textbf{SST-5}} & \multicolumn{1}{c}{\textbf{SNLI}}  & \multicolumn{1}{c}{\textbf{MNLI}} & \multicolumn{1}{c}{\textbf{RTE}} & \multicolumn{1}{c}{\textbf{TREC}} & \multicolumn{1}{c}{\textbf{Average}}\\
    & \multicolumn{2}{c}{------ sentiment ------} & \multicolumn{3}{c}{------ natural language inference ------} & \multicolumn{1}{c}{--- topic ---}\\
    \midrule
    Zero-shot & 79.0  & 35.5  & 50.2  & 48.8  & 51.4  & 32.0 &49.5\\
        LP    &  91.3 ($\pm$0.5) & 51.7 ($\pm$0.5) & 80.9 ($\pm$1.0) & 71.5 ($\pm$1.1) & 73.1 ($\pm$1.5) & 89.4 ($\pm$0.5) &76.3\\
    \midrule
    FT  & 91.9 ($\pm$1.8) & 47.5 ($\pm$1.9) & 77.5 ($\pm$2.6) & 70.0 ($\pm$2.3) & 66.4 ($\pm$7.2) & 85.0 ($\pm$2.5) &73.1\\
    FT (LoRA) & 91.4 ($\pm$1.7) & 46.7 ($\pm$1.1) & 74.9 ($\pm$4.3) & 67.7 ($\pm$1.4) & 66.1 ($\pm$3.5) & 82.7 ($\pm$4.1) &71.6\\
    FT (prefix) &  91.9 ($\pm$1.0) & 47.7 ($\pm$1.1) & 77.2 ($\pm$1.3) & 66.5 ($\pm$2.5) & 66.6 ($\pm$2.0) & 85.7 ($\pm$1.3) &72.6\\
    \midrule
    MeZO  & 93.3 ($\pm$0.7) & 53.2 ($\pm$1.4) & 83.0 ($\pm$1.0) & 78.3 ($\pm$0.5) & 78.6 ($\pm$2.0) & 94.3 ($\pm$1.3) &80.1\\
    % MeZO (LoRA) & 93.4 ($\pm$0.4) & 52.4 ($\pm$0.8) & 84.0 ($\pm$0.8) & 77.9 ($\pm$0.6) & 77.6 ($\pm$1.3) & \textbf{95.0} ($\pm$0.7) &80.1\\
     MeZO (LoRA)& 90.5 ($\pm$0.6) & 45.4($\pm$1.1) & 64.6($\pm$1.2) & 62.1($\pm$0.9) & 61.1($\pm$1.8)  &80.8($\pm$1.5) &67.4 \\
    % \textbf{Reproduce} & 90.5 ($\pm$0.6) & 45.4($\pm$1.1) & 64.6($\pm$1.2) & 62.1($\pm$0.9) & 61.1($\pm$1.8)  &40.8($\pm$1.5) &60.8 \\
    MeZO (prefix) & 93.3 ($\pm$0.1) & 53.6 ($\pm$0.5) & 82.9 ($\pm$1.1) & 75.6 ($\pm$1.2) & 77.2 ($\pm$0.8) & 88.2 ($\pm$0.7) &78.4\\
    \midrule
    HiZOO  & 95.5 ($\pm$0.4) & 53.2 ($\pm$0.9) & 82.6 ($\pm$0.7) & 77.7 ($\pm$0.6) & \textbf{80.0} ($\pm$ 1.5) & \textbf{94.6} ($\pm$1.1) &80.6\\
    HiZOO (LoRA) & 91.7 ($\pm$0.3) & 45.3 ($\pm$0.7) & 76.5 ($\pm$0.3) & 63.1 ($\pm$0.6) & 70.4 ($\pm$1.4) & 85.6 ($\pm$1.5) &72.1\\
    HiZOO (prefix) & \textbf{96.1} ($\pm$0.2) & \textbf{54.2} ($\pm$0.4) & \textbf{85.7} ($\pm$0.7) & \textbf{79.7} ($\pm$1.0) & 77.3 ($\pm$0.2) & 93.9 ($\pm$0.6) &\textbf{81.2}\\
    \bottomrule
    \end{tabular}}
    \vspace{-8pt}
    
    \label{tab:roberta_k512}%
\end{table*}%

\begin{figure}[h]
% \vskip -0.1in
\centering
\scalebox{1.0}{
\includegraphics[width=1.0\linewidth]{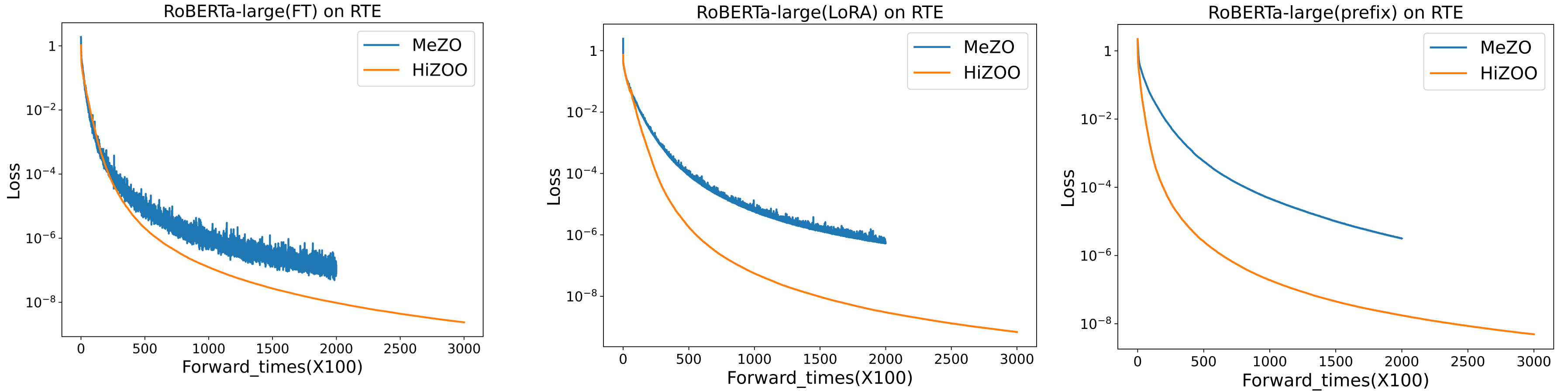}}
\vspace{-5mm}
\caption{Loss curves on RoBERTa-large between MeZO and HiZOO.}
\label{fig:app_roberta1}
%\vskip -0.2in
\end{figure}

\begin{figure}[h]
% \vskip -0.1in
\centering
\scalebox{1.0}{
\includegraphics[width=1.0\linewidth]{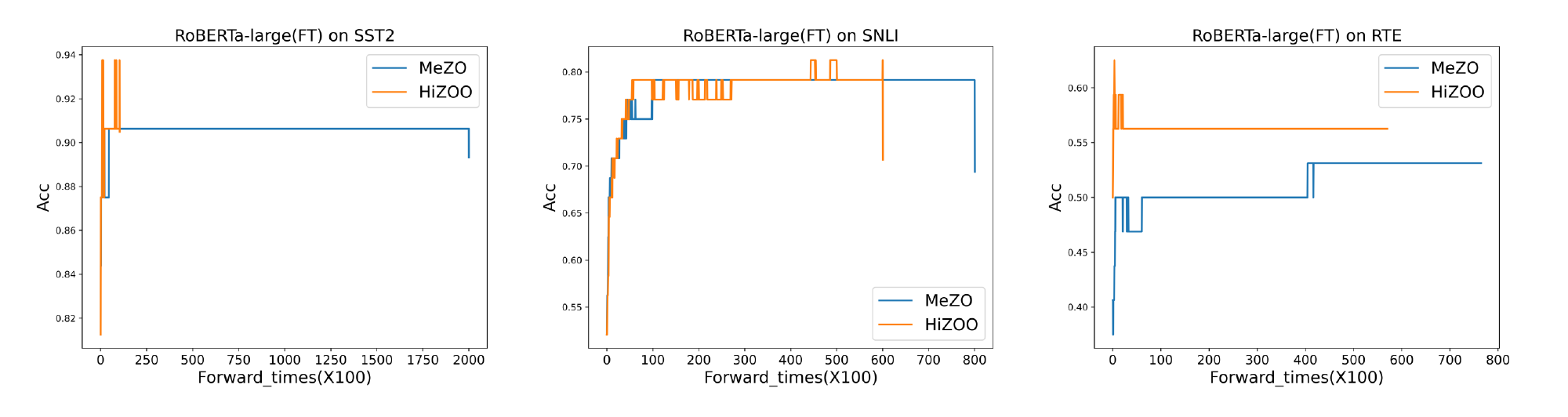}}
\vspace{-5mm}
\caption{Accuracy curves on RoBERTa-large between MeZO and HiZOO.}
\label{fig:app_roberta_eval_acc}
%\vskip -0.2in
\end{figure}

%\subsubsection{Experiment Results}
%\label{app:roberta_result}
In Table \ref{tab:roberta_k512} we show the full experiment results. Additionally, we plot more loss curves to compare with MeZO. As shown in Figure \ref{fig:app_roberta1}, we can see that \nameo can greatly accelerate the training process over MeZO, which verifies the robustness of \name.

% \begin{figure*}[h]
% %\vskip -0.15in
% \centering
% \scalebox{0.99}{
% \includegraphics[width=1\linewidth,trim=0 0 100 205,clip]{appendix/app_roberta1.pdf}}
% \vspace{-5mm}
% \caption{Experiments on RTE with 3 fine-tuning methods.}
% \label{fig:app_roberta2}
% %\vskip -0.15in
% \end{figure*}

% \begin{figure*}[h]
% %\vskip -0.15in
% \centering
% \scalebox{0.99}{
% \includegraphics[width=1\linewidth,trim=0 0 97 205,clip]{appendix/app_roberta2.pdf}}
% \vspace{-5mm}
% \caption{Experiments on SNLI with 3 fine-tuning methods.}
% \label{fig:app_roberta1}
% %\vskip -0.15in
% \end{figure*}

%zyj
\subsection{Detailed results on various LLMs}
\label{app:Llama3_phi2_opt}

\begin{figure}[t]
\centering
    \begin{minipage}{1.0\linewidth}
    \centering
    \scalebox{1.1}{
    \includegraphics[width=0.9\linewidth]{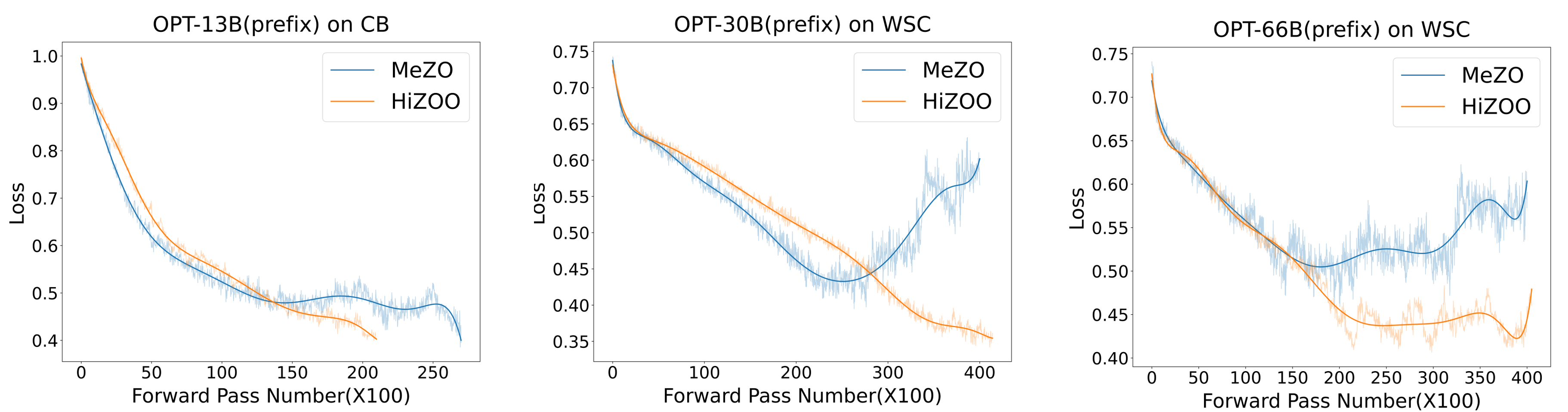}}
    \vspace{-7mm}
    \caption{Loss curves on OPT between MeZO and HiZOO.}
    \vspace{5mm}
    \label{fig:app_opt}
    \end{minipage}
\qquad
    \begin{minipage}{1.0\linewidth}
    \centering
    \scalebox{1.1}{
    \includegraphics[width=0.9\linewidth]{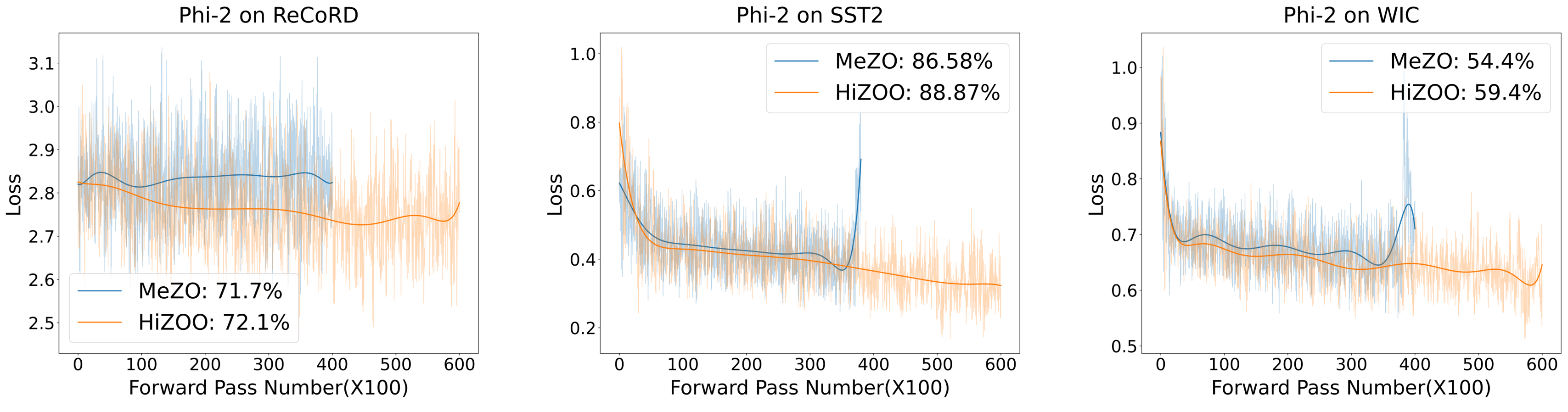}}
    \vspace{-7mm}
    \caption{Loss curves on Phi-2 between MeZO and HiZOO.}
    \vspace{5mm}
    \label{fig:app_Phi2}
    \end{minipage}
\qquad
    \begin{minipage}{1.0\linewidth}
    \centering
    \scalebox{1.1}{
    \includegraphics[width=0.9\linewidth]{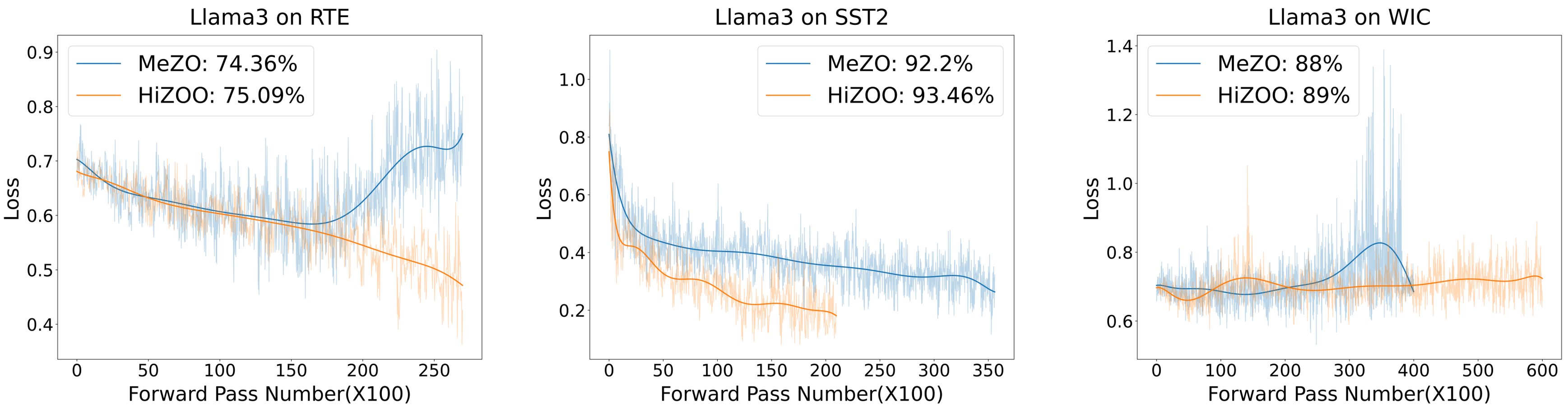}}
    \vspace{-7mm}
    \caption{Loss curves on Llama3 between MeZO and HiZOO.}
    %\vspace{4mm}
    \label{fig:app_Llama3}
    \end{minipage}
\end{figure}

%\subsection{Experiment Hyperparameters}
%\label{app:opt_hyperpara}
We use the hyperparameters in Table~\ref{tab:opt_hyper} for HiZOO experiments on OPT. %Regarding learning rate scheduling and early stopping, we use constant learning rate for all HiZOO experiments. 
Full results for OPT-30B and OPT-66B are in  Table~\ref{tab:app_opt3066}. \textbf{We also provide the relative loss curves of fine-tuning OPT family in Figure~\ref{fig:app_opt}.} We provide several loss curves of fine-tuning Phi-2(2.7B) and Llama3(8B) in Figure~\ref{fig:app_Phi2} and Figure~\ref{fig:app_Llama3}.
% Full results for OPT-13B and OPT-30B are in Table~\ref{tab:opt-13b} and Table~\ref{tab:app_opt3066}. We also provide the relative loss curves of fine-tuning OPT family in Figure~\ref{fig:app_opt}.
\vspace{-3mm}

\begin{table*}[h]
\caption{The hyperparameter grids used for OPT experiments. All weight decay is set to $0$. \nameo uses $20$K steps and constant learning rates.}
    \centering
    \small
    % \resizebox{0.8\columnwidth}{!}{
    \begin{tabular}{lrc}
    \toprule
    Experiment & Hyperparameters & Values \\
    \midrule
    \name & Batch size & $16$ \\
   % & Learning rate & $\{1\mathrm{e}{-6}, 1\mathrm{e}{-7} \}$ or $\{1\mathrm{e}{-6}, 5\mathrm{e}{-7}, 1\mathrm{e}{-7} \}$ for SQuAD and DROP\\
   & Learning rate & $\{1\mathrm{e}{-6}, 5\mathrm{e}{-7}, 1\mathrm{e}{-7} \}$\\
    & $\mu$ & $1\mathrm{e}{-3}$ \\
    \midrule
    \name (prefix) & Batch size & $16$ \\
    & Learning rate & $\{5\mathrm{e}{-2}, 1\mathrm{e}{-2}, 5\mathrm{e}{-3} \}$ \\
    & $\mu$ & $1\mathrm{e}{-1}$ \\
& \# prefix tokens &$5$\\
    \midrule
    \name (LoRA) & Batch size & $16$ \\
    & Learning rate &  $\{1\mathrm{e}{-4}, 5\mathrm{e}{-5}, 1\mathrm{e}{-5} \}$ \\
    & $\mu$ & $1\mathrm{e}{-2}$ \\
    & $(r, \alpha)$ & $(8, 16)$ \\
    \midrule\midrule
    FT with Adam & Batch size & $8$ \\
    & Learning Rates & $\{1\mathrm{e}{-5}, 5\mathrm{e}{-5}, 8\mathrm{e}{-5} \}$\\
    \bottomrule
    \end{tabular}
    % }
    
    \label{tab:opt_hyper}
    \end{table*}
\vspace{-3mm}

\begin{table}[h]
\centering
\vspace{-2mm}
\caption{
         Experiments on OPT-30B and OPT-66B(with $1000$ examples). The best results are highlighted in bold for better comparison. We highlight the best results between HiZOO and MeZO in bold to facilitate comparison.
     }
     %\vspace{-2mm}
\resizebox{0.57\textwidth}{!}{
     \begin{tabular}{lcccccc}
     \toprule
          Task  & \textbf{SST-2} & \textbf{RTE} & \textbf{WSC} & \textbf{WIC} \\ %& \textbf{SQuAD} \\
    \midrule
    30B zero-shot & 56.7 & 52.0 & 38.5 &	50.2 \\ %& 46.5\\
    30B ICL & 81.9 & 66.8  &56.7&	51.3 \\ %& 78.0\\
    30B MeZO & 90.6 & 66.4 & \textbf{63.5} &48.9\\ % & \textbf{85.2}\\
    30B MeZO(prefix) &87.5 & 66.1  &55.8 & 59.1 \\ %&83.9\\

    \midrule
    30B \name &90.3 &\textbf{69.3} & \textbf{63.5} &53.4 \\ %& 62.5\\
    30B \name(prefix) &\textbf{91.2} & 68.6  & 57.7 &\textbf{60.2} \\ % &69.3\\
    \midrule
    \midrule
    
    66B zero-shot & 57.5 &67.2 &	43.3& 50.6 \\ %&48.1\\
    66B ICL & 89.3 & 65.3&	52.9&54.9 \\ %&81.3\\
    66B MeZO(prefix) & \textbf{93.6} & 66.4  & 57.7 &58.6 \\%& 85.0 \\
    
    \midrule
    66B \name(prefix) &\textbf{93.6} & \textbf{71.5}  & \textbf{60.6} &\textbf{61.1} \\%&69.4\\
    
     \bottomrule
     \end{tabular}}
     \vspace{-4mm}
     
     % \vspace{3pt}
     \label{tab:app_opt3066}
 \end{table}

% \begin{table}[h]
% \centering
% \resizebox{0.85\textwidth}{!}{
%      \begin{tabular}{lcccccc}
%      \toprule
%           Task  & \textbf{SST-2} & \textbf{RTE} & \textbf{BoolQ} & \textbf{WSC} & \textbf{WIC} \\ %& \textbf{SQuAD} \\
%     \midrule
%     30B zero-shot & 56.7 & 52.0 & 39.1 & 38.5 &	50.2 \\ %& 46.5\\
%     30B ICL & 81.9 & 66.8 & 66.2 &56.7&	51.3 \\ %& 78.0\\
%     30B MeZO & 90.6 & 66.4 & 67.2 & \textbf{63.5} &48.9\\ % & \textbf{85.2}\\
%     30B MeZO(prefix) &87.5 & 66.1 & \textbf{73.5} &55.8 & 59.1 \\ %&83.9\\

%     \midrule
%     30B \name &90.3 &\textbf{69.3} &70.8 & \textbf{63.5} &53.4 \\ %& 62.5\\
%     30B \name(prefix) &\textbf{91.2} & 68.6 &73.1 & 57.7 &\textbf{60.2} \\ % &69.3\\
%     \midrule
%     \midrule
    
%     66B zero-shot & 57.5 &67.2 & 66.8&	43.3& 50.6 \\ %&48.1\\
%     66B ICL & 89.3 & 65.3&	62.8&	52.9&54.9 \\ %&81.3\\
%     66B MeZO(prefix) & \textbf{93.6} & 66.4 & \textbf{73.7} & 57.7 &58.6 \\%& 85.0 \\
    
%     \midrule
%     66B \name(prefix) &\textbf{93.6} & \textbf{71.5} &73.2 & \textbf{60.6} &\textbf{61.1} \\%&69.4\\
    
%      \bottomrule
%      \end{tabular}}
%      \vspace{4mm}
%      \caption{
%          Experiments on OPT-30B and OPT-66B(with $1000$ examples). The best results are highlighted in bold for better comparison. We highlight the best results between HiZOO and MeZO in bold to facilitate comparison.
%      }
%      % \vspace{3pt}
%      \label{tab:app_opt3066}
%  \end{table}

% \begin{figure}[h]
% %\vspace{2mm}
% \centering
% \scalebox{1.1}{
% \includegraphics[width=0.9\linewidth]{appendix/opt.png}}
% \vspace{-4mm}
% \caption{Loss curves on OPT between MeZO and HiZOO.}
% \label{fig:app_opt}
% %\vskip 0.2in
% \end{figure}

% \begin{figure}[h]
% \vskip -0.1in
% \centering
% \scalebox{1.1}{
% \includegraphics[width=0.9\linewidth]{appendix/phi2.png}}
% \vspace{-5mm}
% \caption{Loss curves on Phi-2 between MeZO and HiZOO.}
% \label{fig:app_Phi2}
% %\vskip -0.2in
% \end{figure}

% \begin{figure}[h]
% \vskip -0.1in
% \centering
% \scalebox{1.1}{
% \includegraphics[width=0.9\linewidth]{appendix/llama3.png}}
% \vspace{-5mm}
% \caption{Loss curves on Llama3 between MeZO and HiZOO.}
% \label{fig:app_Llama3}
% %\vskip -0.2in
% \end{figure}

\section{Details about Memory Usage}% and wall-clock time analysis}
\label{app:memory_time}

Here we show the detailed numbers of memory profiling results Table \ref{tab:app_memory}. We did not turn on any advance memory-saving options, e.g., gradient checkpointing. We set the per-device batch size as 1 to test the minimum hardware requirement to run the model with specific optimization algorithms. We use Nvidia’s $nvidia-smi$ command to monitor the GPU memory usage.

\begin{table*}[h]
  \centering
  \vspace{-4pt}
  \caption{Memory usage on the MultiRC (average tokens=400) dataset. Results of ICL and full-parameter tuning are from MeZO\citep{mezo}.}
  \vspace{2pt}
  \scalebox{0.82}{
    \begin{tabular}{lcccccc}
    \toprule
    &Method & \textbf{zero-shot/MeZO(FT)} & \textbf{\name(FT) } & \textbf{\name-L(FT)} & \textbf{ICL} & \textbf{Adam(FT)}\\
    \midrule
    &1.3B & 1xA100 (4GB) & 1xA100 (7GB) & 1xA100 (4GB) & 1xA100 (6GB)  & 1xA100 (27GB)\\
    &2.7B & 1xA100 (7GB) & 1xA100 (13GB) & 1xA100 (8GB) & 1xA100 (8GB)   & 1xA100 (55GB)\\
    &6.7B & 1xA100 (14GB) & 1xA100 (29GB) & 1xA100 (15GB) & 1xA100 (16GB)  & 2xA100 (156GB)\\
    &13B & 1xA100 (26GB) & 1xA100 (53GB) & 1xA100 (29GB) & 1xA100 (29GB)  & 4xA100 (316GB)\\
    &30B & 1xA100 (58GB) & 2xA100 (118GB) & 1xA100 (64GB) & 1xA100 (62GB)  & 8xA100 (633GB)\\
    &66B & 2xA100 (128GB) & 3xA100 (246GB) & 2xA100 (140GB) & 2xA100 (134GB) & 16xA100\\

    \bottomrule
    \end{tabular}}
    
    \label{tab:app_memory}
    \vspace{-4pt}
\end{table*}%

\section{Details about Wallclock Time Efficiency}
\label{app:train_time}

In this section, we measure the wallclock time efficiency of HiZOO compared to MeZO and full-parameter fine-tuning (FT) with respect to different model sizes. Due to the lack of NV-Link connectivity in our A100 GPUs, we selected models that can be fully fine-tuned on a single A100 GPU for comparison. As shown in Table~\ref{tab:app_train_time}, HiZOO exhibits a longer per-step duration compared to MeZO, within a 50\% margin. This result indicates that the primary overhead in hierarchical optimization methods lies in the forward propagation process. Given that HiZOO involves an additional forward pass compared to MeZO, the time per step increases by approximately 1.4 to 1.5 times.

In conclusion, the speedup factors derived from the forward pass step used in our comparisons between HiZOO and MeZO reflect the actual wallclock time efficiency improvements accurately.

\begin{table*}[htbp]
  \centering
  \vspace{-3pt}
  \caption{Wallclock time per step between MeZO, HiZOO and HiZOO-L. The increase in wallclock time per step for HiZOO compared to MeZO is less than 1.5 times across different model sizes. To avoid introducing additional overheads such as inter-GPU communication, results are measured on the same dataset (SST-2) and GPUs (80GB A100), with each result averaged over 100 steps. "BS" refers to batch size. For the relatively smaller RoBERTa-large model, we used a BS=64, while for models larger than 1B parameters, we used a BS=16.}
  \vspace{4pt}
  \scalebox{1}{
    \begin{tabular}{lcccc}
    \toprule
    Model & \textbf{RoBERTa-large(350M)} & \textbf{Phi-2(2.7B)} & \textbf{Llama3(8B)} & \textbf{OPT(13B)} \\
    \midrule
    MeZO & 0.2092s(BS=64) & 0.3011s(BS=16) & 0.7471s(BS=16) & 1.1108s(BS=16) \\
    HiZOO & 0.3023s(BS=64) & 0.4486s(BS=16) & 1.1090s(BS=16) & 1.5225s(BS=16) \\
    HiZOO-L & 0.3193s(BS=64) & 0.4851s(BS=16) & 1.1996s(BS=16) & 1.6422s(BS=16) \\
    \bottomrule
    \end{tabular}
  }
  
  \label{tab:app_train_time}
\end{table*}

\vspace{-5mm}
\section{Details about Ablation Experiments}
\label{app:ablation}

%\subsection{Influence of Estimation Times $n$ Per Step}
%\label{app:estimation_times}

% We conducted experiments to explore the influence of estimation times $n$ per step as shown in Figure~\ref{fig:app_estimation_times}. We can conclude that when $n$ is larger, the estimation of diagonal Hessian is more accurate, finding the direction of gradient descent more precisely. But it requires more computation with $n$ times per step. So choosing an appropriate value of $n$ is very important during the training. 

% \begin{figure*}[h]
% %\vskip -0.15in
% \centering
% \scalebox{0.80}{
% \includegraphics[width=1\linewidth]{appendix/app_estimate_times.png}}
% %\vspace{-5mm}
% \caption{Influence of estimation times per step. (a) 2D trajectories of gradient descent; (b) Corresponding loss curves.}
% \label{fig:app_estimation_times}
% %\vskip -0.15in
% \end{figure*}
\vspace{-2mm}
\subsection{Influence of Smooth Scale $\alpha_t$ and number of estimation $n$ per step}
\label{app:smooth_scale}
\begin{figure*}[h]
\vspace{-3mm}
\centering
%\scalebox{0.70}{
\includegraphics[width=1\linewidth,trim=0 0 100 205,clip]{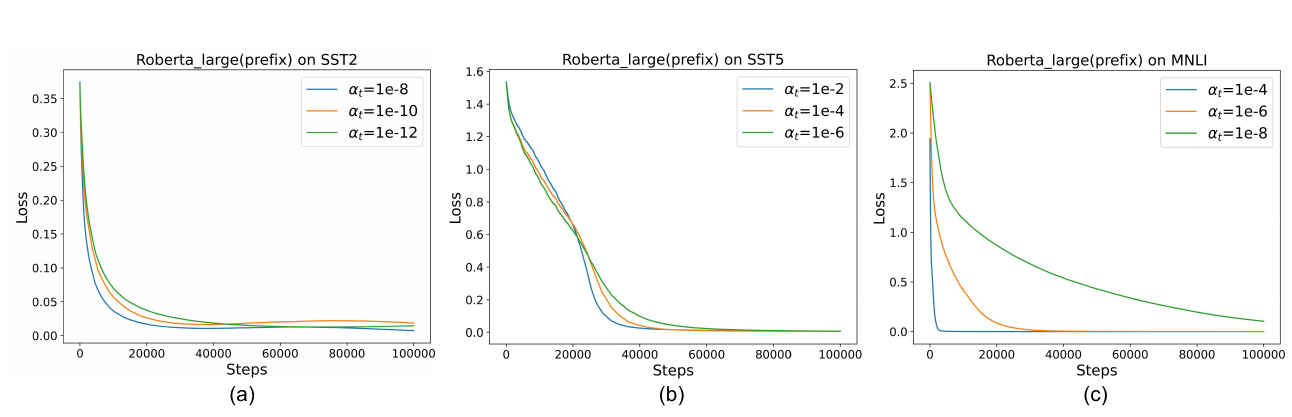}
\vspace{-8mm}
\caption{More experiments on influence of the value of Smooth scale $\alpha_t$ on RoBERTa.}
\label{fig:app_smooth_scale}
%\vskip 0.02in
\end{figure*}

\begin{figure*}[h]
\centering
\scalebox{0.70}{
\includegraphics[width=1\linewidth]{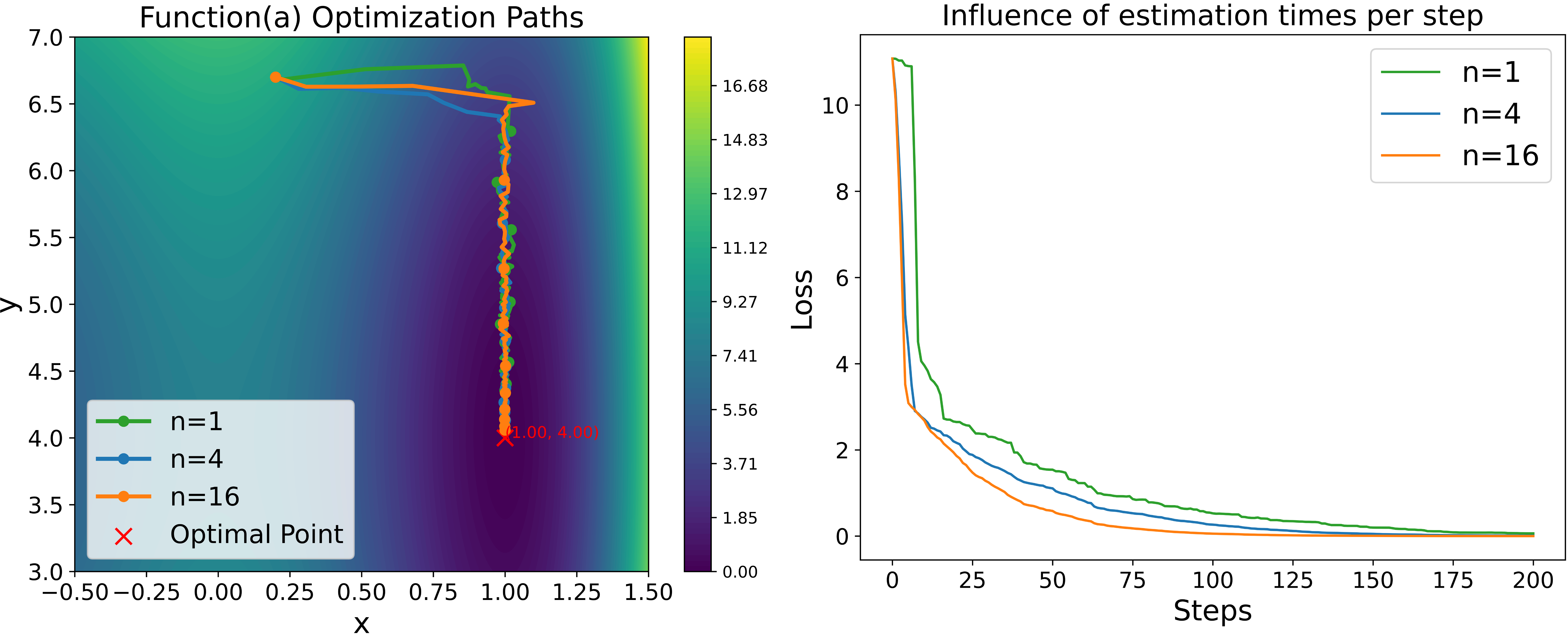}}
\vspace{-2mm}
\caption{Influence of number of estimation per step. (left) 2D trajectories of gradient descent; (right) Corresponding loss curves.}
\label{fig:app_estimation_times}
\end{figure*}

\vspace{-2mm}
We conducted experiments on SST-2, SST-5, MNLI datasets when fine-tuning RoBERTa-large to research the influence of smooth scale $\alpha_t$. Figure~\ref{fig:app_smooth_scale} shows that the value of $\alpha_t$ mainly affects the convergence speed of the model. Additionally, the best value of $\alpha_t$ will vary between different datasets. Figure~\ref{fig:app_estimation_times} shows that the influence of the number of estimation $n$ per steps.

\subsection{Experiments about Omitting $[-\Sigma^{-1}]$ term in Eq.~\eqref{eq:Hessian_update}}
\label{app:error_omitting_term}

\begin{figure*}[h]
%\vskip -0.15in
\centering
% \scalebox{0.70}{
\includegraphics[width=1\linewidth,trim=0 0 100 205,clip]{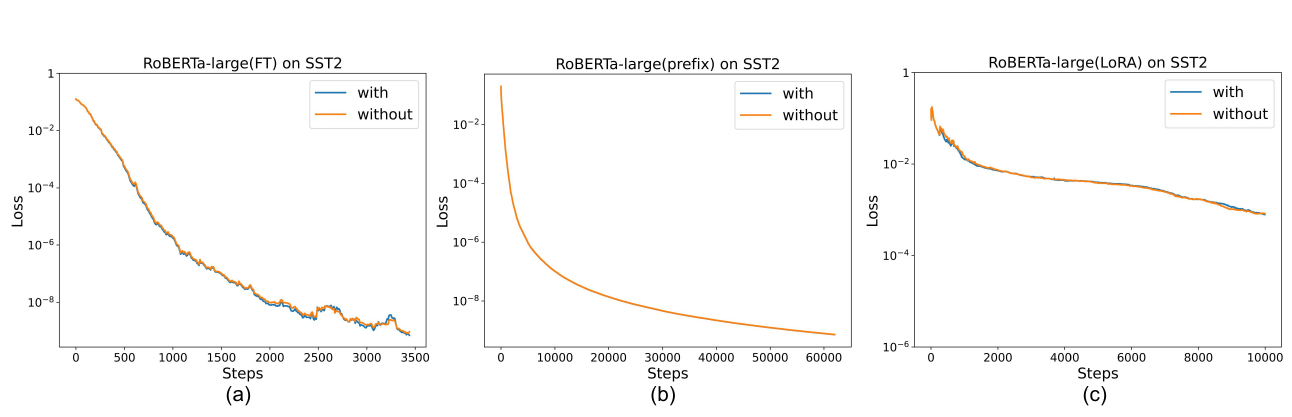}
\vspace{-5mm}
\caption{Experiment about the error generate by omitting the $[-\Sigma^{-1}]$ term in Eq~\eqref{eq:Hessian_update}. 'with' means holding the term and 'without' means omitting the term.}
\label{fig:app_omitting_term}
%\vskip -0.15in
\end{figure*}

We conducted experiments on SST-2 datasets using three methods to fine-tune RoBERTa-large to compare the difference between with $[-\Sigma^{-1}]$ term and without this term. Figure \ref{fig:app_omitting_term} shows that this term can make negligible influence.  

\appendix

\end{document}